\documentclass{article} 
\usepackage{iclr2021_conference,times}

\usepackage{amsmath,amsfonts,bm}

\def\Figref#1{Figure~\ref{#1}}

\def\eqref#1{equation~\ref{#1}}

\def\plaineqref#1{\ref{#1}}

\def\Algref#1{Algorithm~\ref{#1}}

\def\1{\bm{1}}

\DeclareMathAlphabet{\mathsfit}{\encodingdefault}{\sfdefault}{m}{sl}
\SetMathAlphabet{\mathsfit}{bold}{\encodingdefault}{\sfdefault}{bx}{n}

\def\gA{{\mathcal{A}}}

\def\gD{{\mathcal{D}}}

\def\gF{{\mathcal{F}}}

\def\gL{{\mathcal{L}}}

\def\gO{{\mathcal{O}}}

\def\gU{{\mathcal{U}}}

\def\cpe{{\textit{CPE }}}

\newcommand{\R}{\mathbb{R}}

\DeclareMathOperator{\sign}{sign}

\usepackage{hyperref}
\usepackage{url}
\usepackage{graphicx}
\usepackage{algorithm,algorithmicx,algpseudocode}
\usepackage{caption}
\usepackage{subcaption}
\usepackage{amsthm}
\usepackage{paralist}
\usepackage{comment}
\usepackage{enumitem}
\newtheorem{definition}{Definition}
\let\OldStatex\Statex
\renewcommand{\Statex}[1][3]{%
  \setlength\@tempdima{\algorithmicindent}%
  \OldStatex\hskip\dimexpr#1\@tempdima\relax}

\usepackage{booktabs, multirow, multicol}
\usepackage{tabularx}
\newcommand{\tabincell}[2]{\begin{tabular}{@{}#1@{}}#2\end{tabular}}

\usepackage{wrapfig}

\title{
How much progress have we made in neural network training? A New Evaluation Protocol for Benchmarking  Optimizers }

\author{Yuanhao Xiong$^1$, Xuanqing Liu$^1$, Li-Cheng Lan$^1$, Yang You$^2$, Si Si$^3$, Cho-Jui Hsieh$^1$  \\
$^1$Department of Computer Science, UCLA\\
$^2$Department of Computer Science, National University of Singapore \\
$^3$Google Research \\
\texttt{\{yhxiong,xqliu,lclan\}@cs.ucla.edu, youy@comp.nus.edu.sg}\\
\texttt{sisidaisy@google.com, chohsieh@cs.ucla.edu} \\
}

\iclrfinalcopy 
\begin{document}

\maketitle

\begin{abstract}
Many optimizers have been proposed for training deep neural networks, and they often have multiple hyperparameters, which make it tricky to benchmark their performance. 
In this work, we propose a new benchmarking protocol to evaluate both end-to-end efficiency (training a model from scratch without knowing the best hyperparameter) and data-addition training efficiency (the previously selected hyperparameters are used for periodically re-training the model with newly collected data). 
For end-to-end efficiency, unlike previous work that assumes random hyperparameter tuning, which over-emphasizes the tuning time, we propose to evaluate with a bandit hyperparameter tuning strategy. A human study is conducted to show that our evaluation protocol matches human tuning behavior better than the random search. For data-addition training, we propose a new protocol for assessing the hyperparameter sensitivity to data shift. We then apply the proposed benchmarking framework to 7 optimizers and various tasks, including computer vision, natural language processing, reinforcement learning, and graph mining. Our results show that there is no clear winner across all the tasks. 

\end{abstract}

\vspace{-10pt}\section{Introduction}
\vspace{-5pt}
Due to the enormous data size and non-convexity, stochastic optimization algorithms have become widely used in training deep neural networks. In addition to Stochastic Gradient Descent (SGD)~\citep{robbins1951stochastic}, many variations such as Adagrad~\citep{duchi2011adaptive} and Adam~\citep{kingma2014adam} have been proposed. Unlike classical, hyperparameter free optimizers such as gradient descent and Newton's method\footnote{The step sizes of gradient descent and Newton's method can be automatically adjusted by a line search procedure~\citep{nocedal2006numerical}.},  stochastic optimizers often hold multiple hyperparameters including learning rate and momentum coefficients. Those hyperparameters are critical not only to the speed, but also to the final performance, and are often hard to tune.

It is thus non-trivial to benchmark and compare optimizers in deep neural network training. And a biased benchmarking mechanism could lead to a false sense of improvement when developing new optimizers.
In this paper, we aim to rethink the role of hyperparameter tuning in benchmarking optimizers and develop new benchmarking protocols to reflect their performance in practical tasks better. We then benchmark seven recently proposed and widely used optimizers and study their performance on a wide range of tasks. 
In the following, we will first briefly review the two existing benchmarking protocols, discuss their pros and cons, and then introduce our contributions. 

\vspace{-3pt}
\paragraph{Benchmarking performance under the best hyperparameters}
A majority of previous benchmarks and comparisons on optimizers are based on the best hyperparameters. \cite{wilson2017marginal,shah2018minimum} made a  comparison of SGD-based methods against adaptive ones under their best hyperparameter configurations. They found that SGD can outperform adaptive methods on several datasets under careful tuning. Most of the benchmarking frameworks for ML training also assume knowing the best hyperparameters for optimizers~\citep{schneider2019deepobs,coleman2017dawnbench,zhu2018tbd}. Also, the popular MLPerf benchmark evaluated the performance of optimizers under the best hyperparameter. It showed that ImageNet and BERT could be trained in 1 minute using the combination of good optimizers, good hyperparameters, and thousands of accelerators.

Despite each optimizer's peak performance being evaluated, benchmarking under the best hyperparameters makes the comparison between optimizers unreliable and fails to reflect their practical performance. 
First, the assumption of knowing the best hyperparameter is unrealistic. In practice, it requires a lot of tuning efforts to find the best hyperparameter, and the tuning efficiency varies greatly for different optimizers. 
It is also tricky to define the ``best hyperparameter'', which depends on the hyperparameter searching range and grids. Further, since many of these optimizers are sensitive to hyperparameters, some improvements reported for new optimizers may come from insufficient tuning for previous work. 
\vspace{-3pt}
\paragraph{Benchmarking performance with random hyperparameter search}
It has been pointed out in several papers that tuning hyperparameter needs to be considered in evaluating  optimizers~\citep{schneider2019deepobs,asi2019importance}, 
but having a formal evaluation protocol on this topic is non-trivial. 
Only recently, two papers~\citet{choi2019empirical} and \citet{sivaprasad2020optimizer} take hyperparameter tuning time into account when comparing SGD with Adam/Adagrad.
However, their comparisons among optimizers are conducted on random hyperparameter search. We argue that these comparisons could 
over-emphasize the role of hyperparameter tuning, which could lead to a pessimistic and impractical performance benchmarking for optimizers. This is due to the following reasons: 
First, in the random search comparison, each bad hyperparameter has to run fully (e.g., 200 epochs). In practice, a user can always stop the program early for bad hyperparameters if having a limited time budget. For instance, if the learning rate for SGD is too large, a user can easily observe that SGD diverges in a few iterations and directly stops the current job. Therefore, the random search hypothesis will over-emphasize the role of hyperparameter tuning and does not align with a real user's practical efficiency. 
Second, the performance of the best hyperparameter is crucial for many applications. For example, in many real-world applications, we need to re-train the model every day or every week with newly added data. So the best hyperparameter selected in the beginning might benefit all these re-train tasks rather than searching parameters from scratch. In addition, due to the expensive random search, random search based evaluation often focuses on the
 low-accuracy region\footnote{For instance, \cite{sivaprasad2020optimizer} only reaches $<50\%$ accuracy in their CIFAR-100 comparisons.}, while practically we care about the performance for getting reasonably good accuracy. 
 
 \vspace{-3pt}
 \paragraph{Our contributions}
Given that hyperparameter tuning is either under-emphasized (assuming the best hyperparameters) or over-emphasize (assuming random search) in existing benchmarking protocols and comparisons, we develop {\bf new evaluation protocols} to compare optimizers to reflect the real use cases better. Our evaluation framework includes two protocols. First, to evaluate the {\bf end-to-end training efficiency} for a user to train the best model from scratch, we develop an efficient evaluation protocol to compare the accuracy obtained under various time budgets, including the hyperparameter tuning time. Instead of using the random search algorithm, we adopt the Hyperband~\citep{li2017hyperband} algorithm for hyperparameter tuning since it can stop early for bad configurations and better reflect the real running time required by a user. Further, we also propose to evaluate the {\bf data addition training efficiency} for a user to re-train the model with some newly added training data, with the knowledge of the best hyperparameter tuned in the previous training set. 
We also conduct {\bf human studies} to study how machine learning researchers are tuning parameters in optimizers and how that aligns with our proposed protocols. 

Based on the proposed evaluation protocols, we {\bf study how much progress has recently proposed algorithms made compared with SGD or Adam}. 
Note that most of the recent proposed optimizers have shown outperforming SGD and Adam under the best hyperparameters for some particular tasks, but it is not clear whether the improvements are still significant when considering hyper-parameter tuning, and across various tasks. To this end, we conduct comprehensive experiments comparing state-of-the-art training algorithms, including SGD~\citep{robbins1951stochastic}, Adam~\citep{kingma2014adam}, RAdam~\citep{liu2019variance}, Yogi~\citep{zaheer2018adaptive}, LARS~\citep{you2017large}, LAMB~\citep{you2019large}, and Lookahead~\citep{zhang2019lookahead}, on a variety of training tasks including image classification, generated adversarial networks (GANs), sentence classification (BERT fine-tuning), reinforcement learning and graph neural network training.  
Our main conclusions are: 1) On CIFAR-10 and CIFAR-100, all the optimizers including SGD are competitive. 2) Adaptive methods are generally better on more complex tasks (NLP, GCN, RL). 3) There is no clear winner among adaptive methods. Although RAdam is more stable than Adam across  tasks, Adam is still a very competitive baseline even compared with recently proposed methods.

\vspace{-5pt}
\section{Related Work}
\vspace{-5pt}
\paragraph{Optimizers.} 
Properties of deep learning make it natural to apply stochastic first order methods, such as Stochastic Gradient Descent~(SGD)~\citep{robbins1951stochastic}. Severe issues such as a zig-zag training trajectory and a uniform learning rate have been exposed, and researchers have then drawn extensive attention to modify the existing SGD for improvement. Along this line of work, tremendous progresses have been made including SGDM~\citep{qian1999momentum}, Adagrad~\citep{duchi2011adaptive}, RMSProp~\citep{tieleman2012lecture}, and Adam~\citep{kingma2014adam}. These methods utilize  momentums to stabilize and speed up training procedures. Particularly, Adam is regarded as the default algorithm due to its outstanding compatibility. Then variants such as Amsgrad~\citep{reddi2019convergence}, Adabound~\citep{luo2019adaptive}, Yogi~\citep{zaheer2018adaptive}, and RAdam~\citep{liu2019variance} have been proposed to resolve different drawbacks of Adam. Meanwhile, the requirement of large batch training has inspired the development of LARS~\citep{you2017large} and LAMB~\citep{you2019large}. Moreover, \citet{zhang2019lookahead} has put forward a framework called Lookahead to boost optimization performance by iteratively updating two sets of weights.

\vspace{-3pt}
\paragraph{Hyperparameter tuning methods.} 
Random search and grid search~\citep{bergstra2012random} can be a basic hyperparameter tuning method in the literature. However, the inefficiency of these methods stimulates the development of more advanced search strategies. Bayesian optimization methods including \citet{sivaprasad2020optimizer}, \citet{bergstra2011algorithms} and \citet{hutter2011sequential} accelerate random search by fitting a black-box function of hyperparameter and the expected objective to adaptively guide the search direction. Parallel to this line of work, Hyperband~\citep{li2017hyperband} focuses on reducing evaluation cost for each configuration and early terminates relatively worse trials. \citet{falkner2018bohb} proposes BOHB to combine the benefits of both Bayesian Optimization and Hyperband. All these methods still require huge computation resources. A recent work~\citep{metz2020using} has tried to obtain a list of potential hyperparameters by meta-learning from thousands of representative tasks. We strike a balance between effectiveness and computing cost and leverage Hyperband in our evaluation protocol to compare a wider range of optimizers.

\vspace{-5pt}\section{Proposed Evaluation Protocols}
\vspace{-5pt}
In this section, we introduce the proposed evaluation framework for optimizers. We consider two evaluation protocols, each corresponding to an important training scenario: 
\begin{itemize}[noitemsep,topsep=0pt,parsep=0pt,partopsep=0pt,leftmargin=*]
\item {\bf Scenario I (End-to-end training)}: This is the general training scenario, where a user is given an unfamiliar optimizer and task, the goal is to achieve the best validation performance after several trials and errors. In this case, the evaluation needs to include hyperparameter tuning time. We develop an efficiency evaluation protocol to compare the expected performance under different time budgets.
\item {\bf Scenario II (Data-addition training)}: This is another useful scenario encountered in many applications, where the same model needs to be retrained regularly after collecting some fresh data. In this case, a naive solution is to reuse the previously optimal hyperparameters and retrain the model. However, since the distribution is shifted, the result depends on the sensitivity to the distribution shift. 
\end{itemize}
We describe the detailed evaluation protocol for each setting in the following subsections. 
\vspace{-5pt}
\subsection{End-to-end Training Evaluation Protocol}
\vspace{-5pt}
Before introducing our evaluation protocol for Scenario I, we first formally define the concept of optimizer and its hyperparameters.

\begin{definition}
An optimizer is employed to solve a minimization problem $\min_{\theta} \gL(\theta)$ and can be defined by a tuple $o\in\gO = (\gU, \Omega)$, where $\gO$ contains all types of optimizers. $\gU$ is a specific update rule and $\Omega=(\omega_1,\dots, \omega_N)\in\R^N$ represents a vector of $N$ hyperparameters. Search space of these hyperparameters is denoted by $\gF$. Given an initial parameter value $\theta_0$, together with a trajectory of optimization procedure $H_t=\{\theta_s, \gL(\theta_s), \nabla\gL(\theta_s) \}$, the optimizer updates $\theta$ by
\[
\theta_{t+1} = \gU(H_t, \Omega).
\]
\end{definition}
We aim to evaluate the end-to-end time for a user to get the best model, including the hyperparameter tuning time. A recent work~\citep{sivaprasad2020optimizer} assumes that a user conducts random search for finding the best hyperparameter setting. Still, we argue that the random search procedure will {\it over-emphasize} the importance of hyperparameters --- it assumes a user never stops the training even if they observe divergence or bad results in the initial training phase, which is unrealistic. 

\begin{figure}[ht] 
        \vspace{-10pt}
    \centering
    \begin{subfigure}[ht]{0.35\textwidth}
        \centering
        \includegraphics[width=\textwidth, 
        trim={0in 0in 0in 0in},
        clip=false]{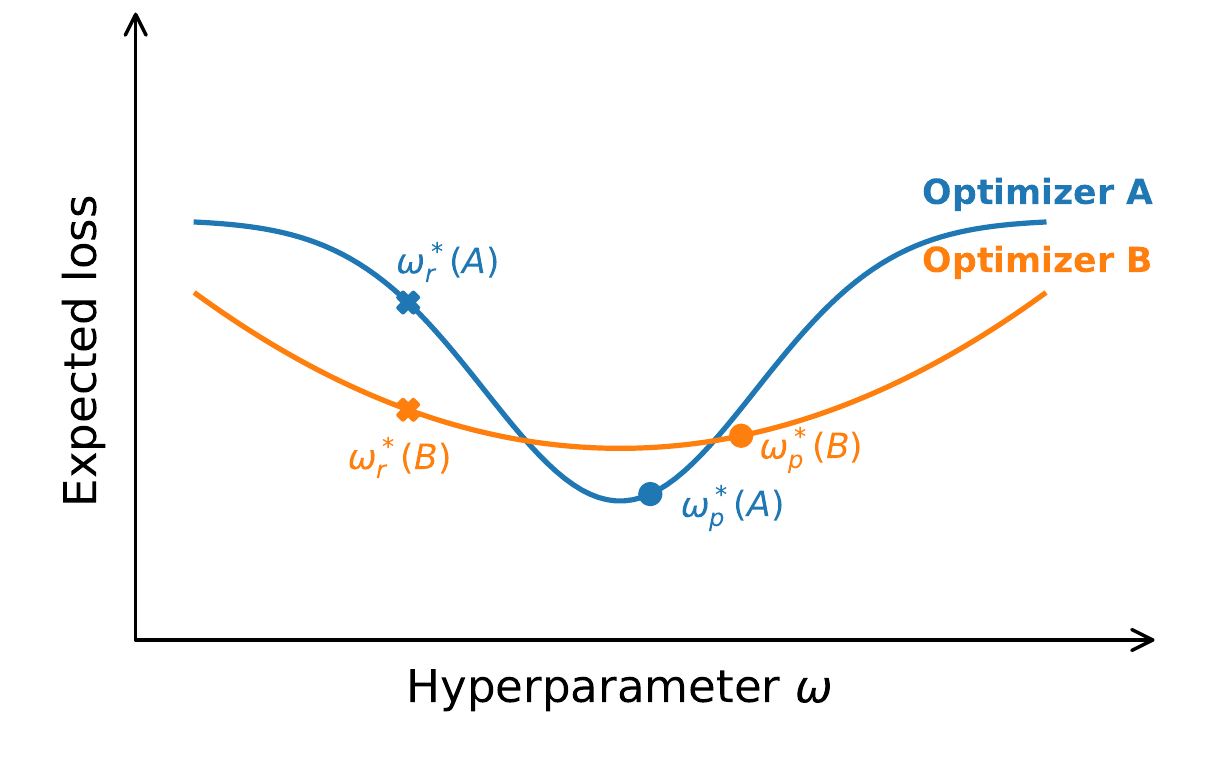}
        \caption{} 
       
        \label{fig:hyper}
    \end{subfigure}
    \hspace{2em}
    \begin{subfigure}[ht]{0.35\textwidth}
    \centering
    \includegraphics[width=\textwidth, 
    trim={0in 0in 0.19in 0in},
    clip=false]{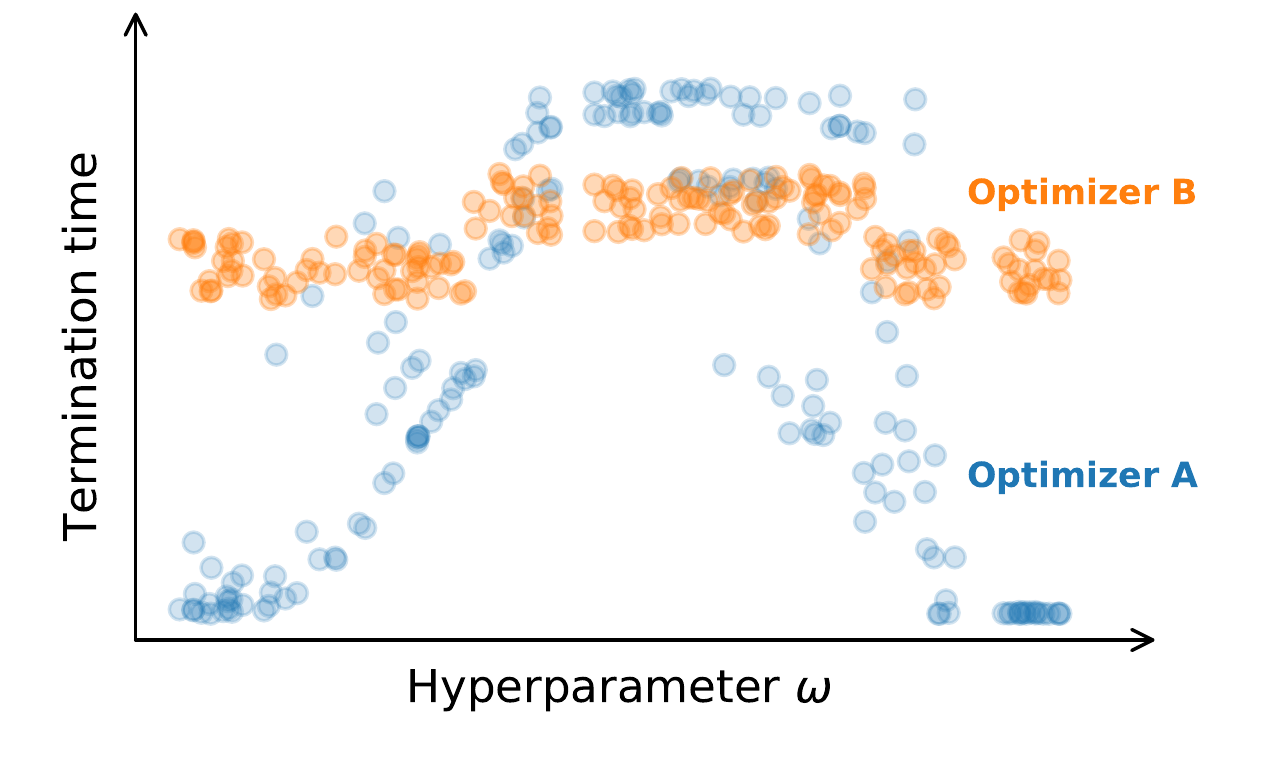}
    \caption{} 
    \label{fig:time}
    \end{subfigure}
    \vspace{-10pt}\caption{An illustration example showing that different hyperparamter tuning methods are likely to affect comparison of optimizers. Optimizer A is more sensitive to hyperparamters than optimizers B, but it may be prefered if bad hyperparameters can be terminated in the early stage. }
    \label{fig:illustration}
    \vspace{-5pt}
\end{figure}

\Figref{fig:illustration} illustrates why random search might not lead to a fair comparison of optimizers. In \Figref{fig:illustration}, we are given two optimizers, A and B, and their corresponding loss w.r.t. hyperparameter. According to \citet{sivaprasad2020optimizer}, optimizer B is considered better than optimizer A since most regions of the hyperparameter space of A outperforms B. This statement is 
reasonable under a random search with a constrained budget. For instance, suppose we randomly sample the same hyperparamter setting for A and B. The final config $\omega_r^*(B)$ found under this strategy can have a lower expected loss than that of $\omega_r^*(A)$, as shown in Figure~\ref{fig:hyper}. However, there exists a more practical search strategy which can invalidate this statement: a user can early terminate a configuration trial when trapped in bad results or diverging. Hence, we can observe in Figure~\ref{fig:time} that for optimizer A, this strategy early-stops many configurations and only allow a limited number of trials to explore to the deeper stage. Therefore, the bad hyperparameters will not affect the overall efficiency of Optimizer A too much. 
In contrast, for optimizer B, performances of different hyperparameters are relatively satisfactory and hard to distinguish, resulting in similar and long termination time for each trial. Therefore,  it may be easier for a practical search strategy $p$ to find the best configuration $\omega_p^*(A)$ of optimizer A than $\omega_p^*(B)$, given the same constrained budget.

This example suggests that random search may over-emphasize the parameter sensitivity when benchmarking optimizers. To better reflect a practical hyperparameter tuning scenario, our evaluation assumes a user applies {\bf Hyperband}~\citep{li2017hyperband}, a simple but effective hyperparameter tuning scheme to get the best model. Hyperband formulates hyperparameter optimization as a unique bandit problem. It accelerates random search through adaptive resource allocation and early-stopping, as demonstrated in Figure \ref{fig:time}. Compared with its more complicated counterparts such as BOHB~\citep{falkner2018bohb}, Hyperband requires less computing resources and performs similarly within a constrained budget. The algorithm is presented in Appendix \ref{appendix: hyperband}.

 \begin{wrapfigure}{r}{0.45\textwidth} 
     \centering
     \vspace{-15pt}     \includegraphics[width=0.45\textwidth]{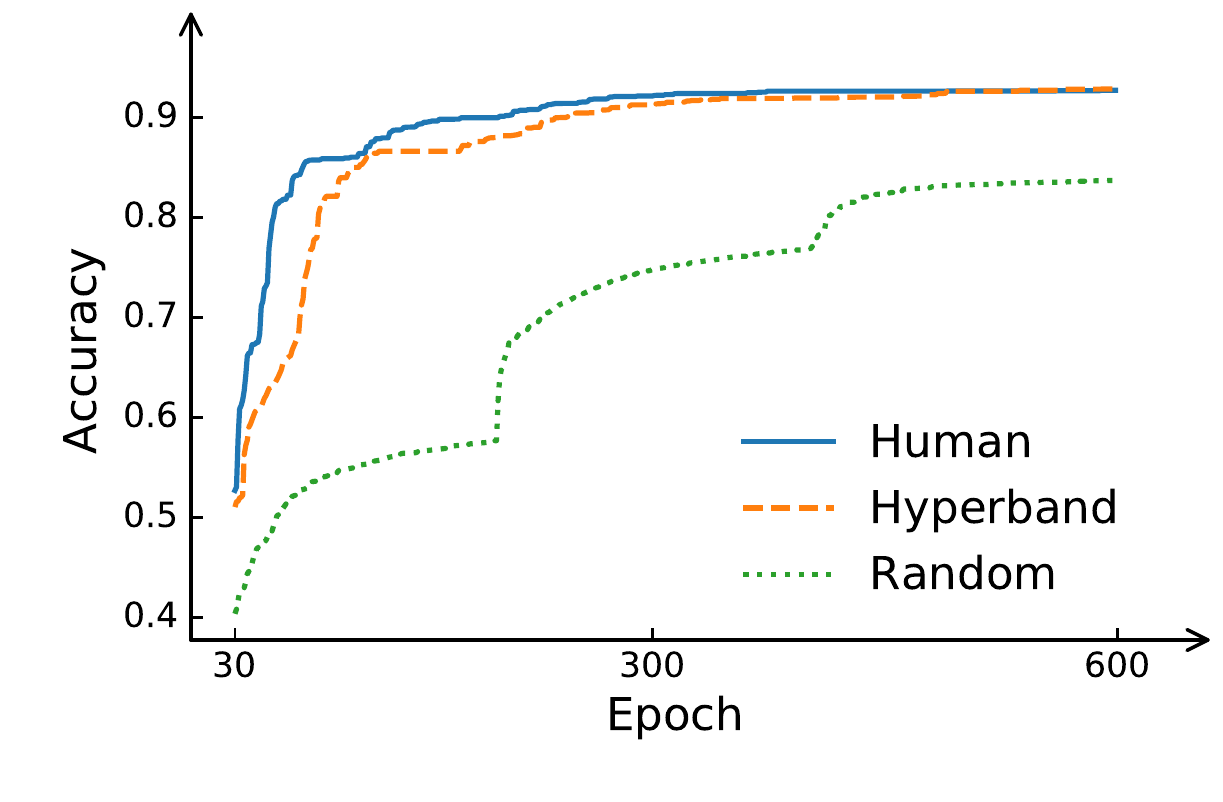}
 \vspace{-25pt}    \caption{Hyperband tuning  used in our evaluation protocol is closer to human behavior than random search. \label{fig:human}}
 \vspace{-6pt}
     \end{wrapfigure}
To verify that Hyperband tends to better capture the human tuning behavior than random search, we conduct a human study as follows: for image classification on CIFAR10, given $10$ learning rate configurations of SGD in the grid $[1.0\times 10^{-8}, 1.0\times 10^{-7}, 1.0\times 10^{-6},\dots, 10]$, participants are requested to search the best one at their discretion. Namely, they can stop or pause a trial any time and continue to evaluate a new configuration until they feel it has already reached the best performance. We collect several data points and average them as human performance. In Figure~\ref{fig:human}, we plot curves for hyperparameter tuning of human, Hyperband, and random search. We find that Hyperband matches humans' behavior better, while random search tends to trap in suboptimal configurations.
This finding justified the use of Hyperband in optimizer benchmarking.

With Hyperband incorporated in end-to-end training, we assume that each configuration is run sequentially and record the best performance obtained at time step $t$ as $P_t$. $\{P_t\}_{t=1}^{T}$ forms a trajectory for plotting learning curves on test set like Figure \ref{fig:cifar10}. Although it is intuitive to observe the performance of different optimizers according to such figures, summarizing a learning curve into a quantifiable, scalar value can be more insightful for evaluation. Thus, as shown in Eq. \plaineqref{eq:tunability}, we use $\lambda$-tunability defined in \cite{sivaprasad2020optimizer} to further measure the performance of optimizers:
\begin{equation}
\label{eq:tunability}
\lambda\text{-tunability}= \sum\nolimits_{t=1}^{T} \lambda_t \cdot P_t, \text{ where}\sum\nolimits_{t}\lambda_t=1 \text{ and } \forall_t \lambda_t >0.
\end{equation}
 One intuitive way is to set $\lambda_t=\1_{t=T}$ to determine which optimizer can reach the best model performance after the whole training procedure. However, merely considering the peak performance is not a good guidance on the choice of optimizers. In practice, we tend to take into account the complete trajectory and exert more emphasis on the early stage. Thus, we employ the Cumulative Performance-Early weighting scheme where $\lambda_t \propto (T-i)$, to compute $\lambda$-tunablity instead of the extreme assignment $\lambda_t=\1_{t=T}$. The value obtained is termed as \textit{CPE} for simplicity.

We present our evaluation protocol in \Algref{al:efficiency}. As we can see, end-to-end training with hyperparameter optimization is conducted for various optimizers on the given task. The trajectory $\{P_t\}_{t=1}^{T}$ is recorded to compute the peak performance as well as \cpe value. Note that the procedure is repeated several times to obtain a reliable result.
\begin{algorithm}[ht]
    \caption{End-to-End Efficiency Evaluation Protocol}
    \label{al:efficiency}
    \textbf{Input:} A set of optimizers $\gO = \{o: o=(\gU, \Omega)\}$, task $a\in\gA$, feasible search space $\gF$ 
    
    \begin{algorithmic}[1]
    \For{$o\in\gO$}
    \For{$i = 1$ \textbf{to} $M$}
    \State Conduct hyperparameter search in $\gF$ with the optimizer $o$ using HyperBand on $a$
    \State Record the performance trajectory $\{P_t\}_{t=1}^{T}$ explored by HyperBand
    \State Calculate the peak performance and \textit{CPE} by Eq.~\plaineqref{eq:tunability} 
    \EndFor
    \State Average peak and \cpe values over $M$ repetitions for the optimizer $o$
    \EndFor
    \State Evaluate optimizers according to their peak and \cpe values
    \end{algorithmic}
\end{algorithm}





\vspace{-10pt}
\subsection{Data-addition Training Evaluation Protocol}
In Scenario II, 
we have a service (e.g., a search or recommendation engine) and we want to re-train the model every day or every week with some newly added training data.
One may argue that an online learning algorithm should be used in this case, but in practice online learning is unstable and industries still prefer this periodically retraining scheme which is more stable. 


In this scenario, once the best hyperparameters were chosen in the beginning, we can reuse them for every training, so no hyperparameter tuning is required and the performance (including both efficiency and test accuracy) under the best hyperparameter becomes important. However, an implicit assumption made in this process is that {\it ``the best hyperparameter will still work when the training task slightly changes''}. This can be viewed as transferability of hyperparameters for a particular optimizer, and our second evaluation protocol aims to evaluate this practical scenario. 

We simulate data-addition training with all classification tasks, and the evaluation protocol works as follows: 1) Extract a subset $\gD_\delta$ containing partial training data from the original full dataset $\gD$ with a small ratio $\delta$; 2) Conduct a hyperparameter search on $\gD_\delta$ to find the best setting under this scenario; 3) Use these hyperparameters to train the model on the complete dataset; 4) Observe the potential change of the ranking of various optimizers before and after data addition.
For step 4) when comparing different optimizers, we will plot the training curve in the full-data training stage (e.g., see Figure~\ref{fig:cifar_addition}), and also summarize the training curve using the \cpe value. The detailed evaluation protocol is described in Algorithm~\ref{al:data-addition}.


\begin{algorithm}[ht]
    \caption{Data-Addition Training Evaluation Protocol}
    \label{al:data-addition}
    
    \textbf{Input:} A set of optimizers $\gO = \{o: o=(\gU, \Omega)\}$, task $a\in\gA$ with a full dataset $\gD$, a split ratio $\delta$ 
    
    \begin{algorithmic}[1]
    \For{$o \in \gO$}
    \For{$i = 1$ \textbf{to} $M$}
    \State Conduct hyperparameter search with the optimizer $o$ using Hyperband on $a$ with a partial dataset $\gD_\delta$, and record the best hyperparameter setting $\Omega_{\text{partial}}$ found under this scenario
    \State Apply the optimizer with $\Omega_{\text{partial}}$ on $\gD_\delta$ and $\gD$, then save the training curves
    \EndFor
    \State Average training curves of $o$ over $M$ repetitions to compute \cpe
    \EndFor
    \State Compare performance of different optimizers under data-addition training
    \end{algorithmic}
\end{algorithm}
\section{Experimental Results}

\paragraph{Optimizers to be evaluated.} As shown in Table \ref{tab:optimizers}, we consider 7 optimizers including non-adaptive methods using only the first-order momentum, and adaptive methods considering both first-order and second-order momentum. We also provide lists of tunable hyperparameters for different optimizers in Table~\ref{tab:optimizers}. Moreover, we consider following two combinations of tunable hyperparameters to better investigate the performance of different optimizers: \textbf{a)} only tuning initial learning rate with the others set to default values and \textbf{b)} tuning a full list of hyperparameters. A detailed description of optimizers as well as default values and search range of these hyperparameters can be found in Appendix \ref{appendix:details}. Note that we adopt a unified search space for a fair comparison following~\citet{metz2020using}, to eliminate biases of specific ranges for different optimizers.
The tuning budget of Hyperband is determined by three items: maximum resource (in this paper we use epoch) per configuration $R$, reduction factor $\eta$, and number of configurations $n_c$. According to \citet{li2017hyperband}, a single Hyperband execution contains $n_s=\lfloor \log_{\eta}(R)\rfloor+1$ of SuccessiveHalving, each referred to as a bracket. These brackets take strategies from least to most aggressive early-stopping, and each one is designed to use approximately $B=R\cdot n_s$ resources, leading to a finite total budget. The number of randomly sampled configurations in one Hyperband run is also fixed and grows with $R$. Then given $R$ and $\eta$, $n_c$ determines the repetition times of Hyperband. We set $\eta=3$ as this default value performs consistently well, and $R$ to a value which each task usually takes for a complete run. For $n_c$, it is assigned as what is required for a single Hyperband execution for all tasks, except for BERT fine-tuning, where a larger number of configurations is necessary due to a relatively small $R$. In Appendix \ref{appendix:details}, we give assigned values of $R$, $\eta$, and $n_c$ for each task.


\begin{table}[ht]
    \centering
    \resizebox{.6\textwidth}{!}{
    \begin{tabular}{c|cc}
    \toprule
      \multicolumn{2}{c}{\textbf{Optimizer}}   & \textbf{Tunable hyperparameter} \\
      \midrule
    \multirow{2}*{Non-adptive} & SGD   & $\alpha_0$, $\mu$ \\
    &  LARS & $\alpha_0$, $\mu$, $\epsilon$ \\
      \midrule
 Adaptive &    \tabincell{c}{Adam, RAdam, Yogi\\ Lookahead, LAMB}  & $\alpha_0$, $\beta_1$, $\beta_2$, $\epsilon$ \\
     \bottomrule
    \end{tabular}
    }
    \caption{Optimizers to be evaluated with their tunable hyperparameters. Specifically, $\alpha_0$ represents the initial learning rate. $\mu$ is the decay factor of the first-order momentum for non-adaptive methods while $\beta_1$ and $\beta_2$ are coefficients to compute the running averages of first-order and second-order momentums. $\epsilon$ is  a small scalar used to prevent division by $0$.}
    \label{tab:optimizers}
    \vspace{-10pt}
\end{table}

\paragraph{Tasks for benchmarking.} For a comprehensive and reliable assessment of optimizers, we consider a wide range of tasks in different domains. When evaluating end-to-end training efficiency, 
we implement our protocol on tasks covering several popular and promising applications in Table \ref{tab:tasks}. Apart from common tasks in computer vision and natural language processing, we introduce two extra tasks in graph neural network training and reinforcement learning. For simplicity, we will use the dataset to represent each task in our subsequent tables of experimental results. (For the reinforcement learning task, we just use the environment name.) The detailed settings and parameters for each task can be found in Appendix \ref{appendix:task}. 

\begin{table}[ht]
    \centering
        \resizebox{.8\textwidth}{!}{
    \begin{tabular}{ccccc}
    \toprule
        \multicolumn{1}{c}{\bf Domain} & {\bf Task} & {\bf Metric} & {\bf Model} & {\bf Dataset} \\
    \midrule
    \multirow{3}*{Computer Vision}     & \tabincell{c}{Image\\Classification} & Accuracy & ResNet-50 & \tabincell{c}{CIFAR10\\CIFAR100} \\
     & VAE & NLL & CNN Autoencoder & CelebA \\
     & GAN & FID & SNGAN network & CIFAR10 \\
     \midrule
     NLP & GLUE benchmark & Accuracy & RoBERTa-base & MRPC \\
     \midrule
     Graph network training & Node labeling & F1 score & Cluster-GCN & PPI \\
     \midrule
     Reinforcement Learning & Walker2d-v3 & Reward & PPO & $\times$ \\
     \bottomrule
    \end{tabular}
    }
    \caption{Tasks for benchmarking optimizers. Details are provided in Appendix~\ref{appendix:task}.}
    \label{tab:tasks}
    \vspace{-10pt}
\end{table}
\subsection{End-to-end efficiency (Secnario I)}
\label{sec:s1}
\begin{figure}[ht] 
    \centering
    \begin{subfigure}[ht]{0.245\textwidth}
        \centering
        \includegraphics[width=\textwidth, 
        trim={0in 0in 0in 0in},
        clip=false]{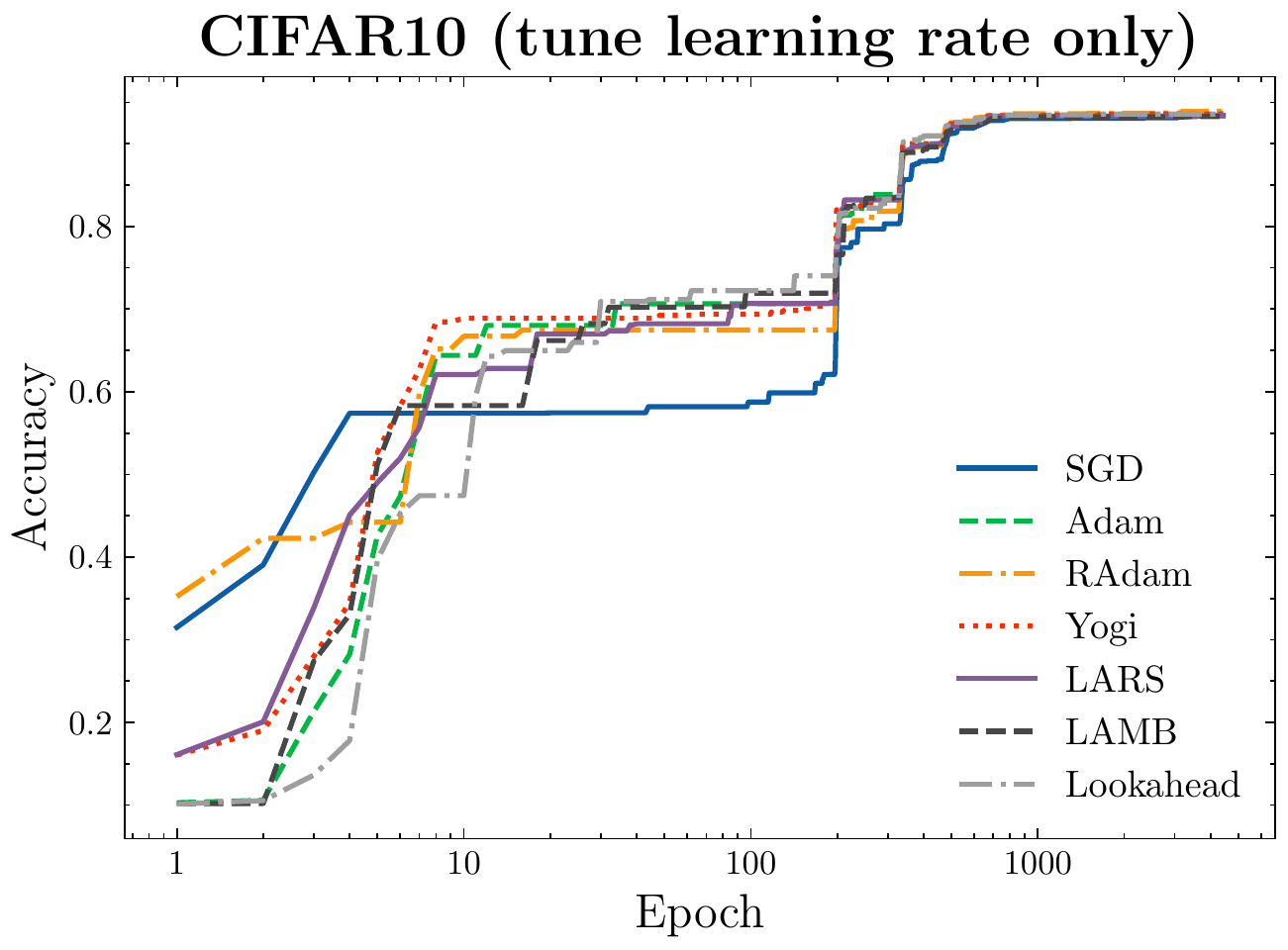}
        \caption{} 
        \label{fig:1param}
    \end{subfigure}
    \begin{subfigure}[ht]{0.245\textwidth}
    \centering
    \includegraphics[width=\textwidth, 
    trim={0in 0in 0in 0in},
    clip=false]{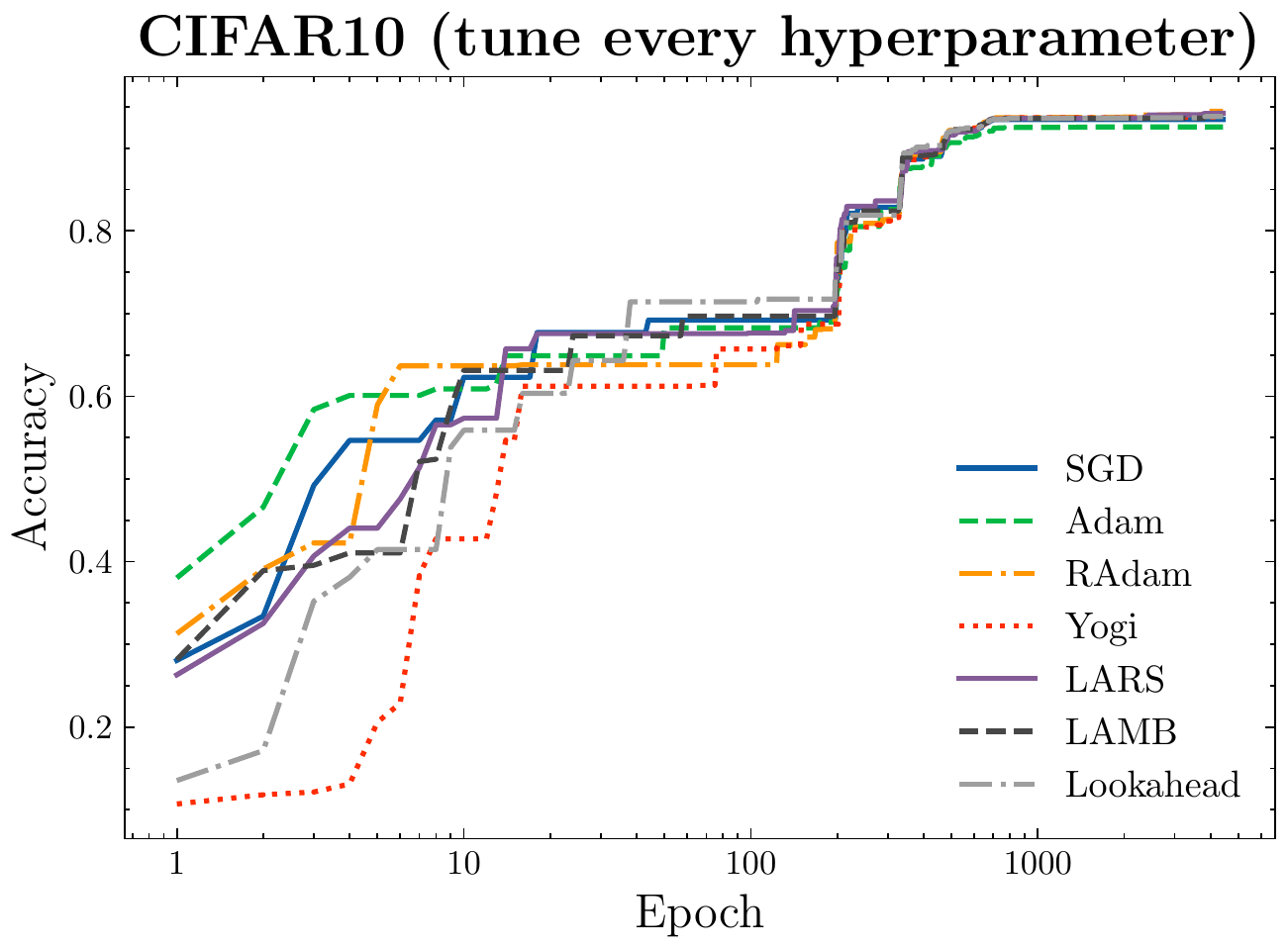}
    \caption{} 
    \label{fig:full}
    \end{subfigure}
        \centering
    \begin{subfigure}[ht]{0.245\textwidth}
        \centering
        \includegraphics[width=\textwidth, 
        trim={0in 0in 0in 0in},
        clip=false]{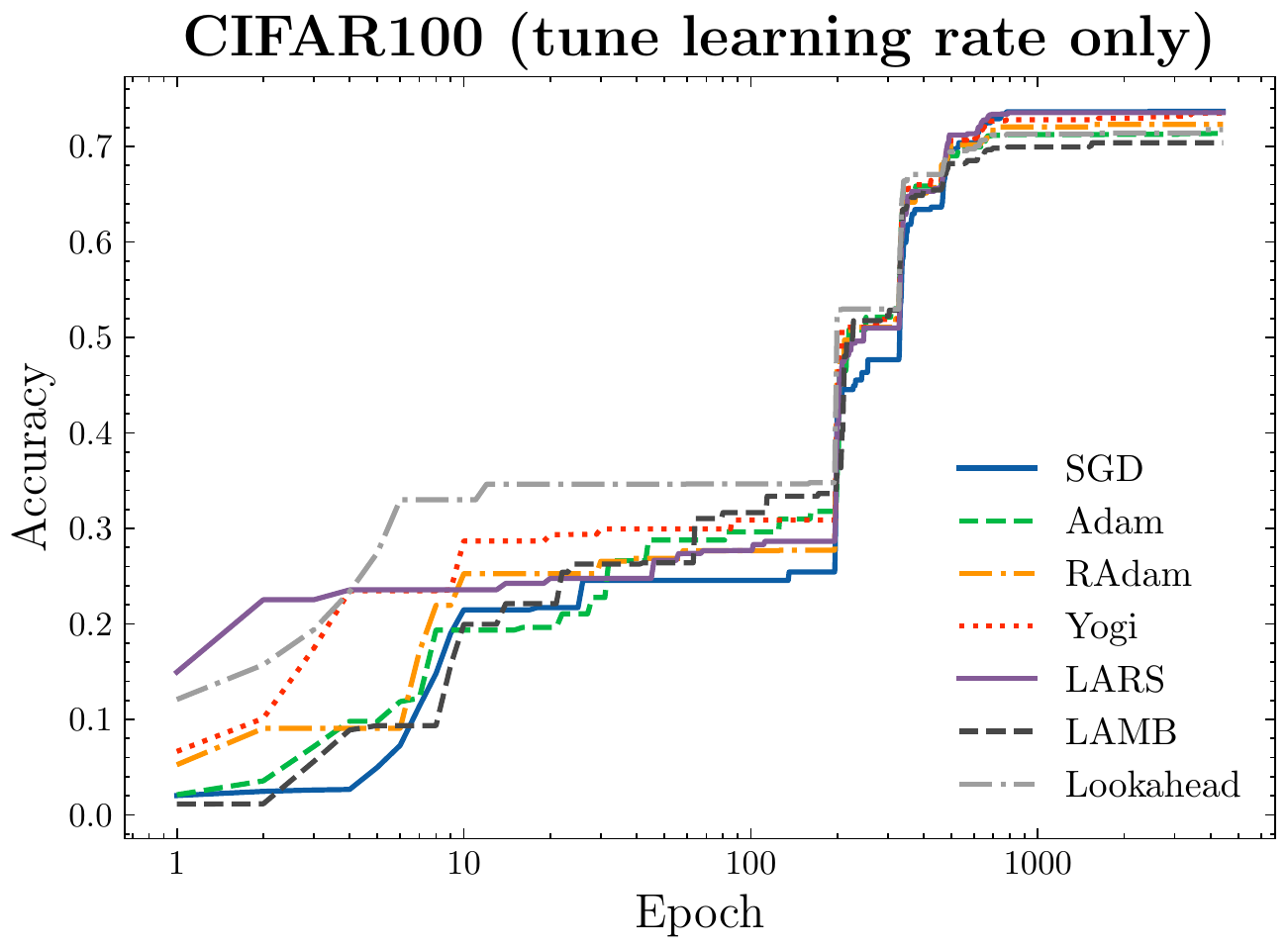}
        \caption{} 
        \label{fig:cifar100-1param-decay}
    \end{subfigure}
    \begin{subfigure}[ht]{0.245\textwidth}
    \centering
    \includegraphics[width=\textwidth, 
    trim={0in 0in 0in 0in},
    clip=false]{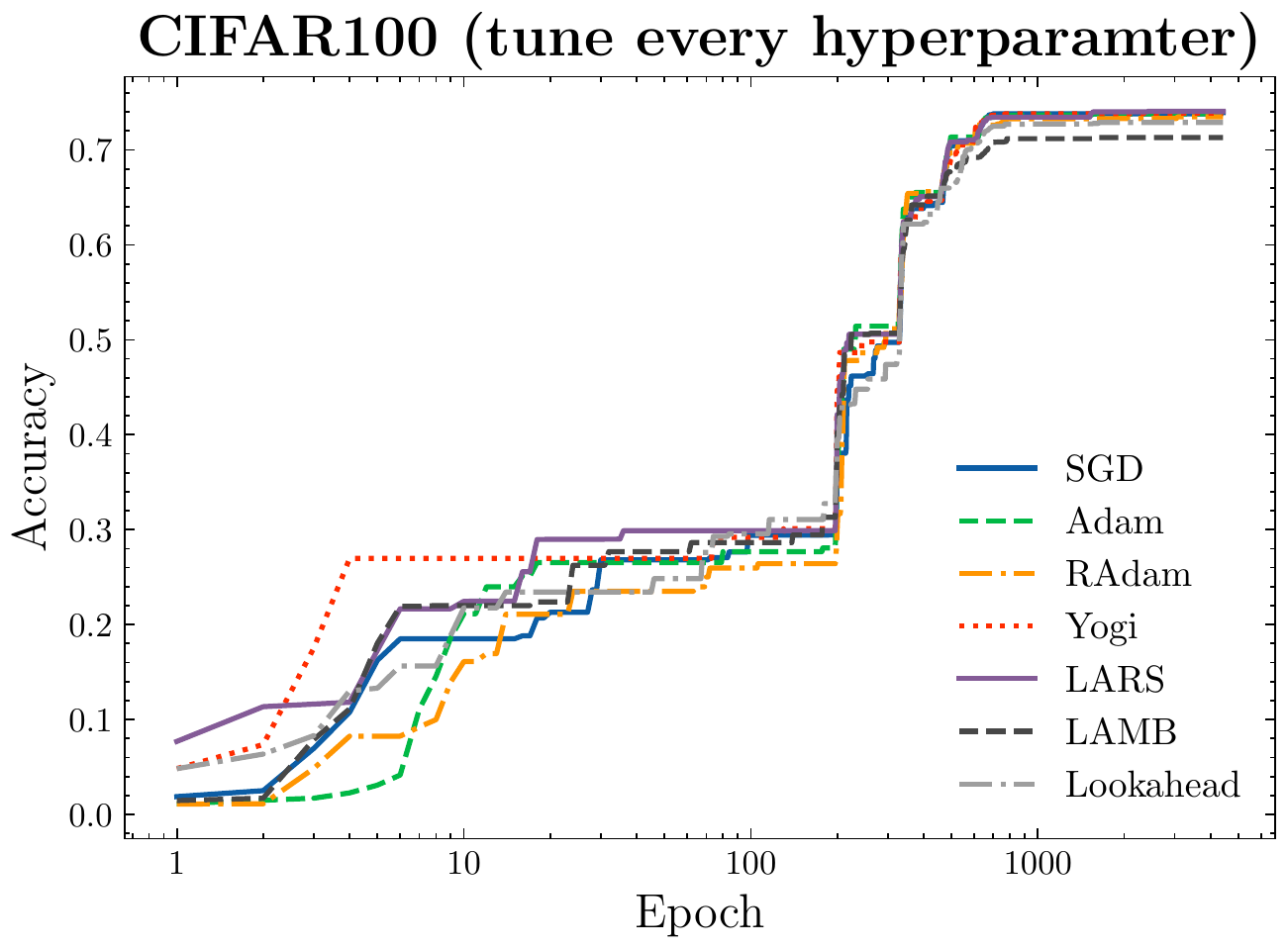}
    \caption{} 
    \label{fig:cifar100-full-decay}
    \end{subfigure}
    \vspace{-10pt}
    \caption{End-to-end training curves with Hyperband on CIFAR10 and CIFAR100. }
    \label{fig:cifar10}
\end{figure}

\begin{figure}[ht] 
    \centering
    \begin{subfigure}[ht]{0.4\textwidth}
        \centering
        \includegraphics[width=\textwidth, 
        trim={0in 0in 0in 0in},
        clip=false]{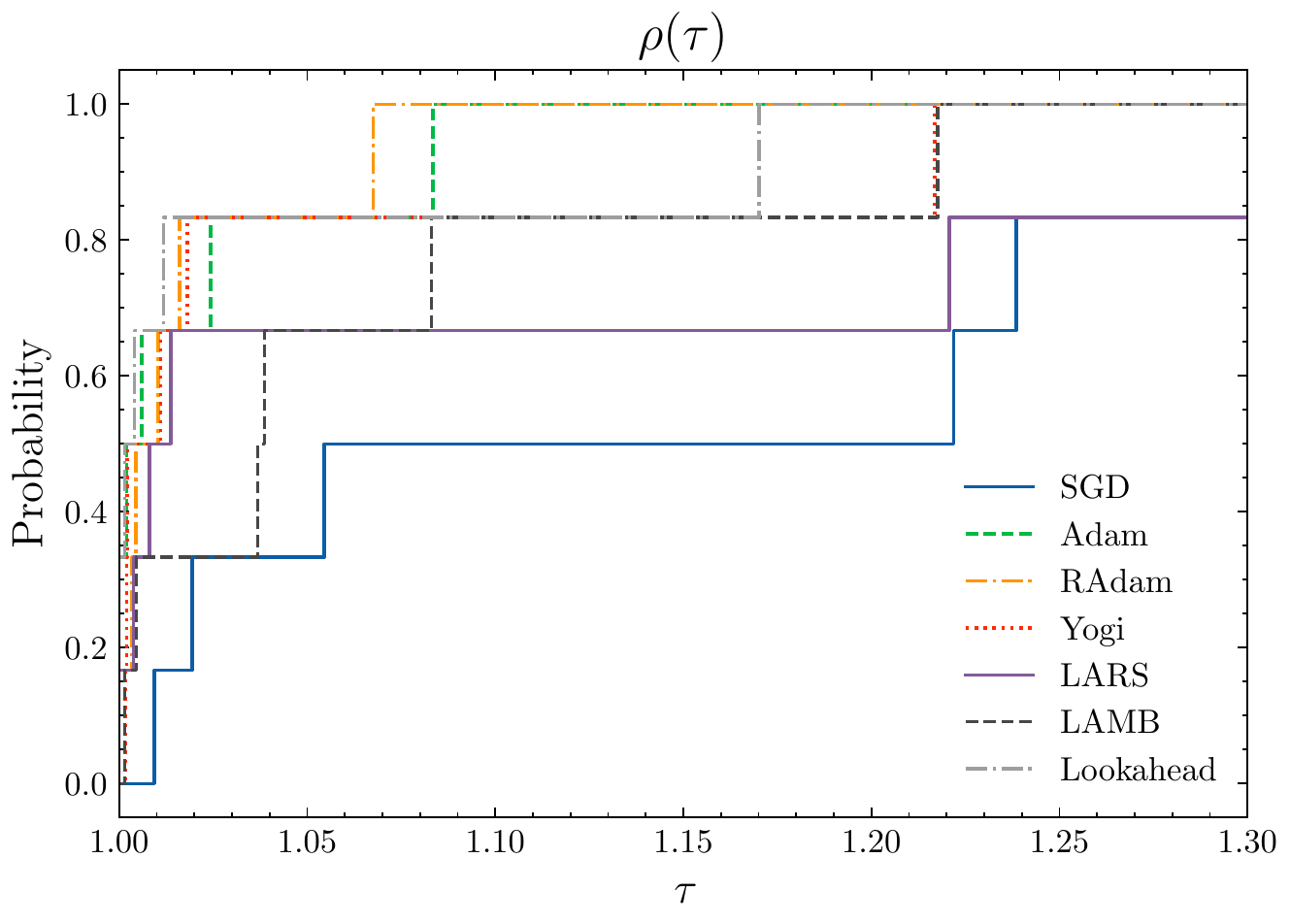}
        \caption{Tune learning rate only.} 
    \end{subfigure}
    \hspace{2em}
    \begin{subfigure}[ht]{0.4\textwidth}
    \centering
    \includegraphics[width=\textwidth, 
    trim={0in 0in 0in 0in},
    clip=false]{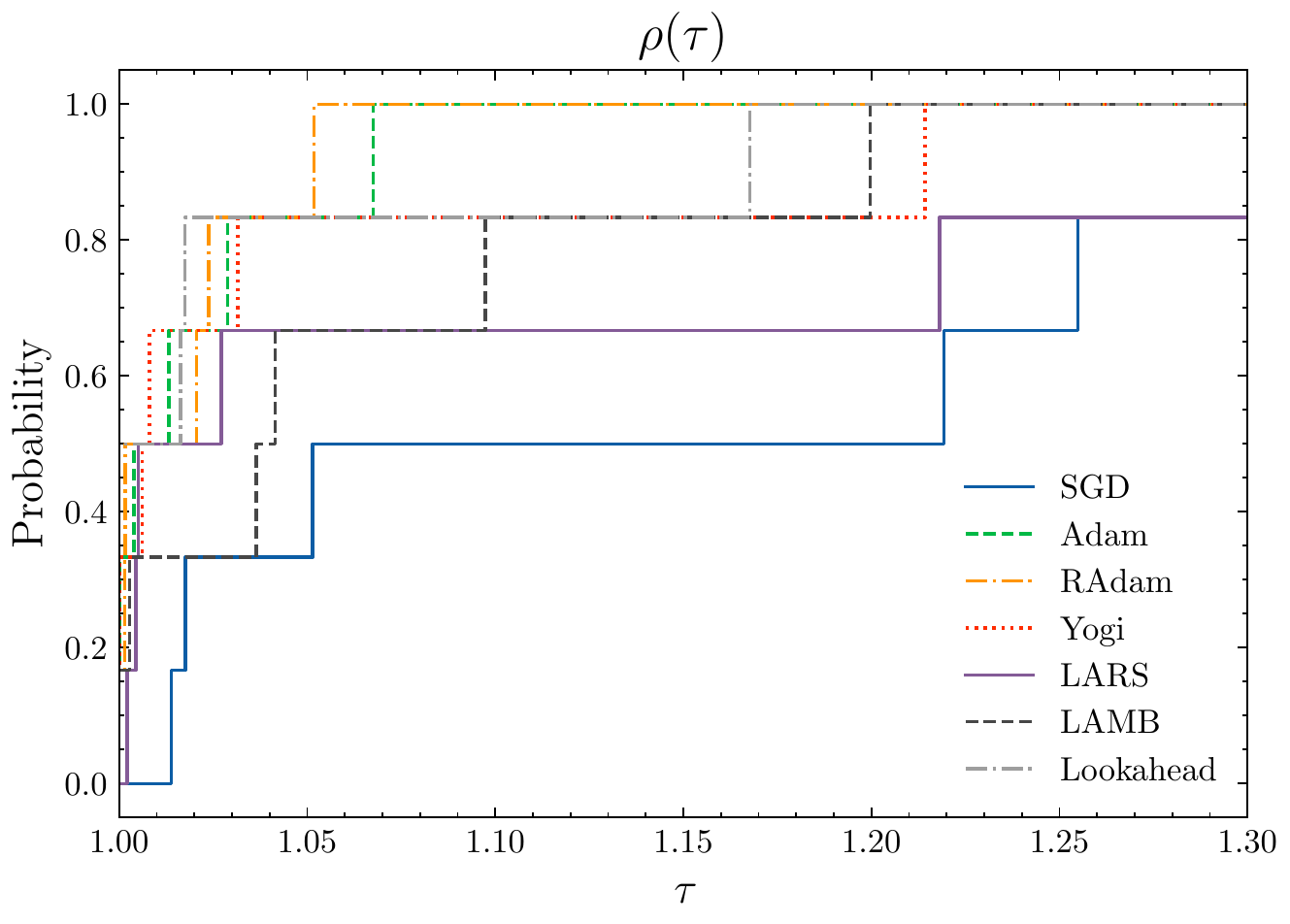}
    \caption{Tune every hyperparameter.} 
    \end{subfigure}
    \vspace{-4pt}
    \caption{Performance profile of $7$ optimizers in the range $[1.0, 1.3]$.}
    \vspace{-15pt}
    \label{fig:rho}
\end{figure}

To evaluate end-to-end training efficiency, we adopt the protocol in Algorithm~\ref{al:efficiency}. Specifically, we record the average training trajectory with Hyperband $\{P_t\}_{t=1}^{T}$ for each optimizer on benchmarking tasks, where $P_t$ is the evaluation metric for each task (e.g., accuracy, reward). We visualize these trajectories in Figure \ref{fig:cifar10} for CIFAR10 and CIFAR100, and calculate \cpe and peak performance in Table~\ref{tab:cpe} and \ref{tab:peak} respectively. More results for other tasks and the peak performance can be found in Appendix \ref{appendix:results}. Besides, in Eq.~\ref{eq:ratio} we compute \textit{performance ratio} $r_{o,a}$ for each optimizer and each task, and then utilize the distribution function of a performance metric called \textit{performance profile} $\rho_{o}(\tau)$ to summarize the performance of different optimizers over all the tasks. For tasks where a lower \cpe is better, we just use $r_{o,a}=\textit{CPE}_{o,a}/\min\{\cpe_{o,a}\}$ instead to guarantee $r_{o,a}\geq1$. The function $\rho_o(\tau)$ for all optimizers is presented in Figure \ref{fig:rho}. Based on the definition of performance profile~\citep{dolan2002benchmarking}, the optimizers with large probability $\rho_{o}(\tau)$ are to be preferred. In particular, the value of $\rho_o(1)$ is the probability that one optimizer will win over the rest and can be a reference for selecting the proper optimizer for an unknown task.
\begin{equation}
    \label{eq:ratio}
    r_{o,a} = \frac{\max\{ \textit{CPE}_{o,a}: o \in \gO\}}{\textit{CPE}_{o,a}}, \quad     \rho_o(\tau) = \frac{1}{|\gA|}\text{size}\big\{a\in\gA: r_{o,a}\leq \tau \big\}.
    \vspace{-10pt}
\end{equation}

Our findings are summarized below: 
\begin{itemize}[noitemsep,topsep=0pt,parsep=0pt,partopsep=0pt,leftmargin=*]
\item In \cite{sivaprasad2020optimizer}, a random hyperparameter search is used for evaluating the efficiency of optimizers, and they conclude that Adam usually outperforms SGD. However, it can be observed from Table~\ref{tab:cpe} and \ref{tab:peak}, that under our protocol based on Hyperband, SGD performs similarly to Adam in terms of  efficiency as well as peak performance, and can even surpass it in some cases like training on CIFAR100. This is mainly because Hyperband can early-stop the runs for bad configurations and thus they will affect less to the final performance.   
\item For image classification tasks all the methods are competitive, while adaptive methods tend to perform better in more complicated tasks (NLP, GCN, RL). 
\item There is no significant distinction among adaptive variants.
Performance of adaptive optimizers tends to fall in the range within $1\%$ of the best result.
\item According to performance profile  in Figure \ref{fig:rho}, the RAdam achieves probability 1 with the smallest $\tau$, and
Adam is the second method achieving that. 
This indicates that RAdam and Adam are achieving relatively stable and consistent performance among these tasks. 


\end{itemize}

\begin{table}[ht]
\caption{CPE for different optimizers on benchmarking tasks. The best performance is highlighted in bold and blue and results within the $1\%$ range of the best are emphasized in bold only.
}
\vspace{-10pt}
\label{tab:cpe}
\begin{center}
\scalebox{0.9}{
   \resizebox{.9\textwidth}{!}{
\begin{tabular}{ccccccc}
\toprule
\multicolumn{1}{c}{\bf Optimizer}  &\multicolumn{1}{c}{\bf CIFAR10 (\%) $\uparrow$} &\multicolumn{1}{c}{\bf CIFAR100 (\%) $\uparrow$}
&\multicolumn{1}{c}{\bf CelebA $\downarrow$}
&\multicolumn{1}{c}{\bf MRPC $\uparrow$}
&\multicolumn{1}{c}{\bf PPI (\%) $\uparrow$}
&\multicolumn{1}{c}{\bf Walker2d-v3 $\uparrow$}
\\ \midrule
\multicolumn{7}{l}{\textit{Tune learning rate only:}}\\
SGD & 88.87  & \textbf{66.85}   & 0.1430 & 69.90 & 76.77  & 2795 \\
Adam & \textbf{90.42}  & 65.88 & \color{blue}\textbf{0.1356}   & \textbf{84.90}  &  \color{blue}\textbf{95.08}  & 3822\\
RAdam & \textbf{90.29}  & 66.41  & \textbf{0.1362}  & \color{blue}\textbf{85.41}  & 94.10  & 3879\\
Yogi & \textbf{90.42}  & \textbf{67.37}  & 0.1371 & 70.19 & 93.39  & \textbf{4132} \\
LARS & \textbf{90.25}  & \color{blue}\textbf{67.48}   & \textbf{0.1367}    & 69.97 & 93.79  & 2986\\
LAMB & \textbf{90.19} & 65.08  & \textbf{0.1358} & 82.23 & 87.79  & 3401\\
Lookahead & \color{blue}\textbf{90.60}  & 65.60 & \textbf{0.1358}   & 72.99 & \textbf{94.69}   & \color{blue}\textbf{4141} \\
\midrule
\multicolumn{7}{l}{\textit{Tune every hyperparameter:}}\\
SGD & \textbf{90.20}  & \textbf{67.36}  & 0.1407 & 71.53 & 94.64& 2978\\
Adam & 89.27  & \textbf{67.57}   & 0.1389  & \color{blue}\textbf{85.23}  & 92.62 & \color{blue}\textbf{4080} \\
RAdam & \textbf{90.14}   & 66.90 & \textbf{0.1366}  & 84.32 & 93.05 & 3813\\
Yogi & \textbf{89.83}  & \textbf{67.65}  & 0.1401 & 68.42 & 88.94  & 3778\\
LARS & \textbf{90.42}   & \color{blue}\textbf{67.78}  & 0.1375 & 77.40  & \color{blue}\textbf{96.34}  & 2728\\
LAMB & \textbf{90.27}   & 65.59  & 0.1382  & \textbf{84.66}  & 93.18  & 2935\\
Lookahead & \color{blue}\textbf{90.44}   & 66.46 & \color{blue}\textbf{0.1360}   & 79.05 & 94.30  & 3786\\
\bottomrule
\end{tabular}
}
}
\end{center}
\vspace{-10pt}
\end{table}


\subsection{Data-addition Training (Scenario II)}
\label{sec:s2}
\begin{figure}[ht] 
    \centering
    \begin{subfigure}[ht]{0.245\textwidth}
        \centering
        \includegraphics[width=\textwidth,
        trim={0in 0in 0in 0in},
        clip=false]{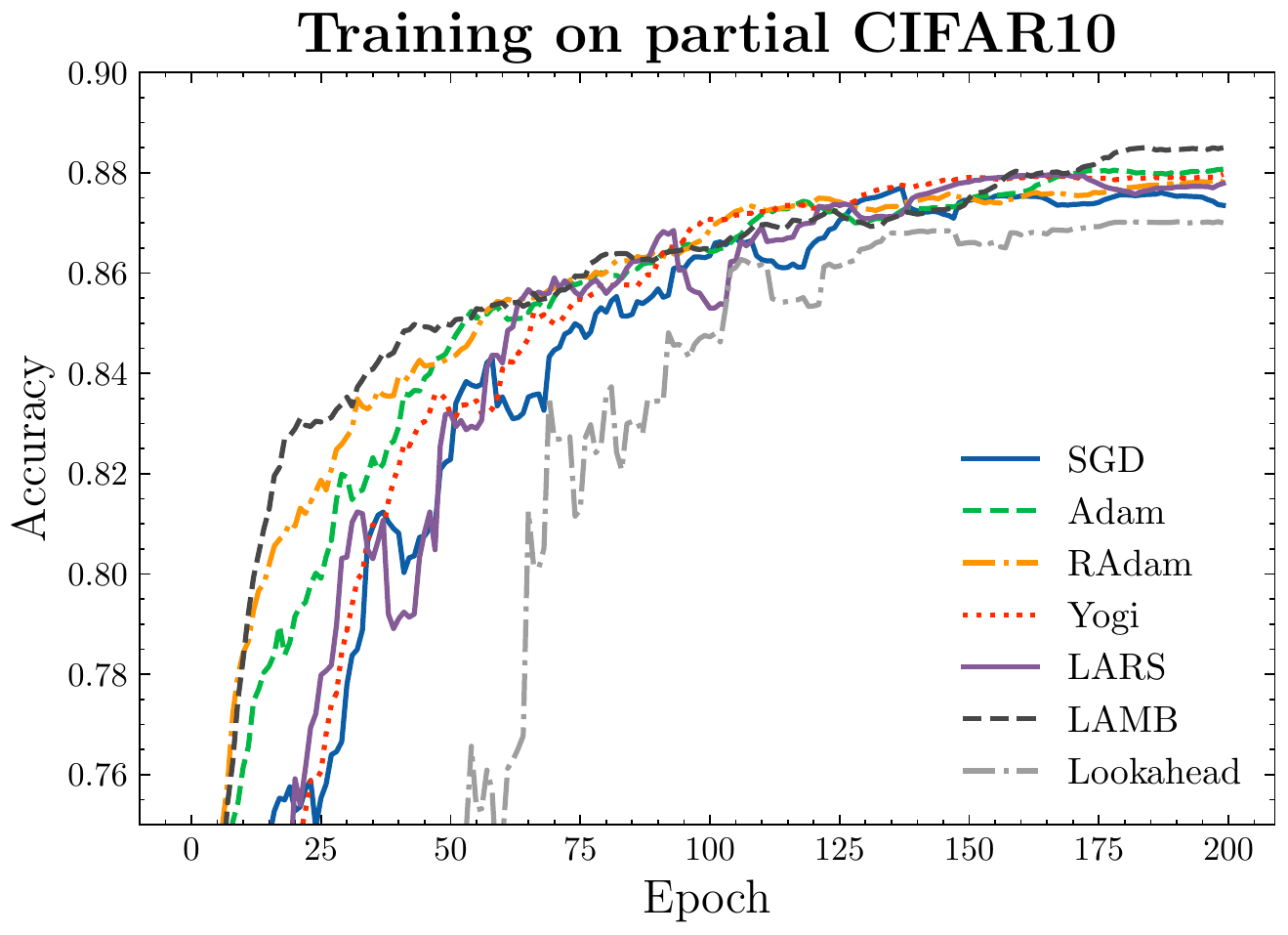}
        \caption{} 
    \end{subfigure}
    \begin{subfigure}[ht]{0.245\textwidth}
        \centering
        \includegraphics[width=\textwidth, 
        trim={0in 0in 0in 0in},
        clip=false]{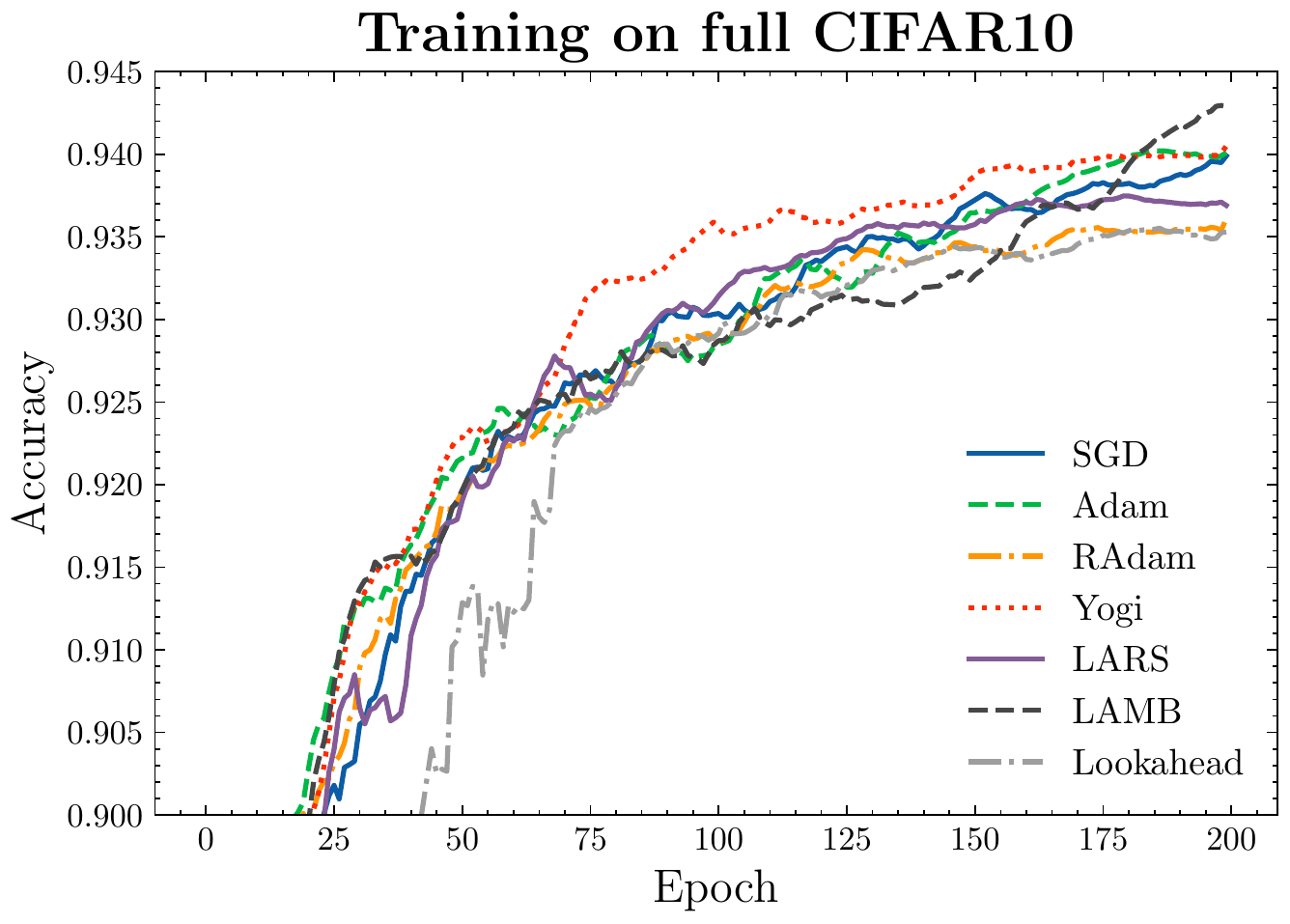}
        \caption{} 
        \label{fig:cifar10-addition}
    \end{subfigure}
    \centering
    \begin{subfigure}[ht]{0.245\textwidth}
        \centering
        \includegraphics[width=\textwidth,
        trim={0in 0in 0in 0in},
        clip=false]{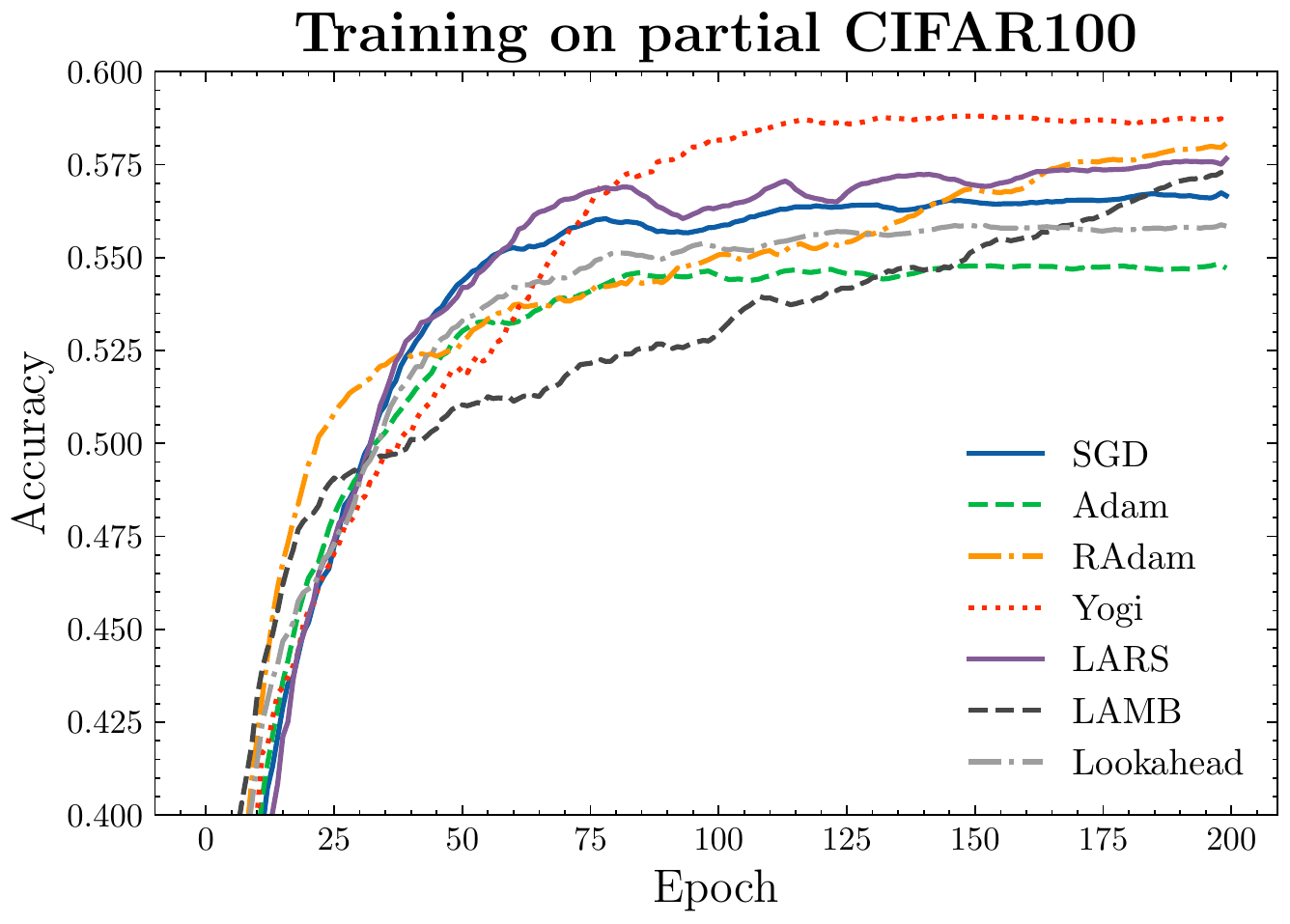}
        \caption{} 
    \end{subfigure}
    \begin{subfigure}[ht]{0.245\textwidth}
        \centering
        \includegraphics[width=\textwidth, 
        trim={0in 0in 0in 0in},
        clip=false]{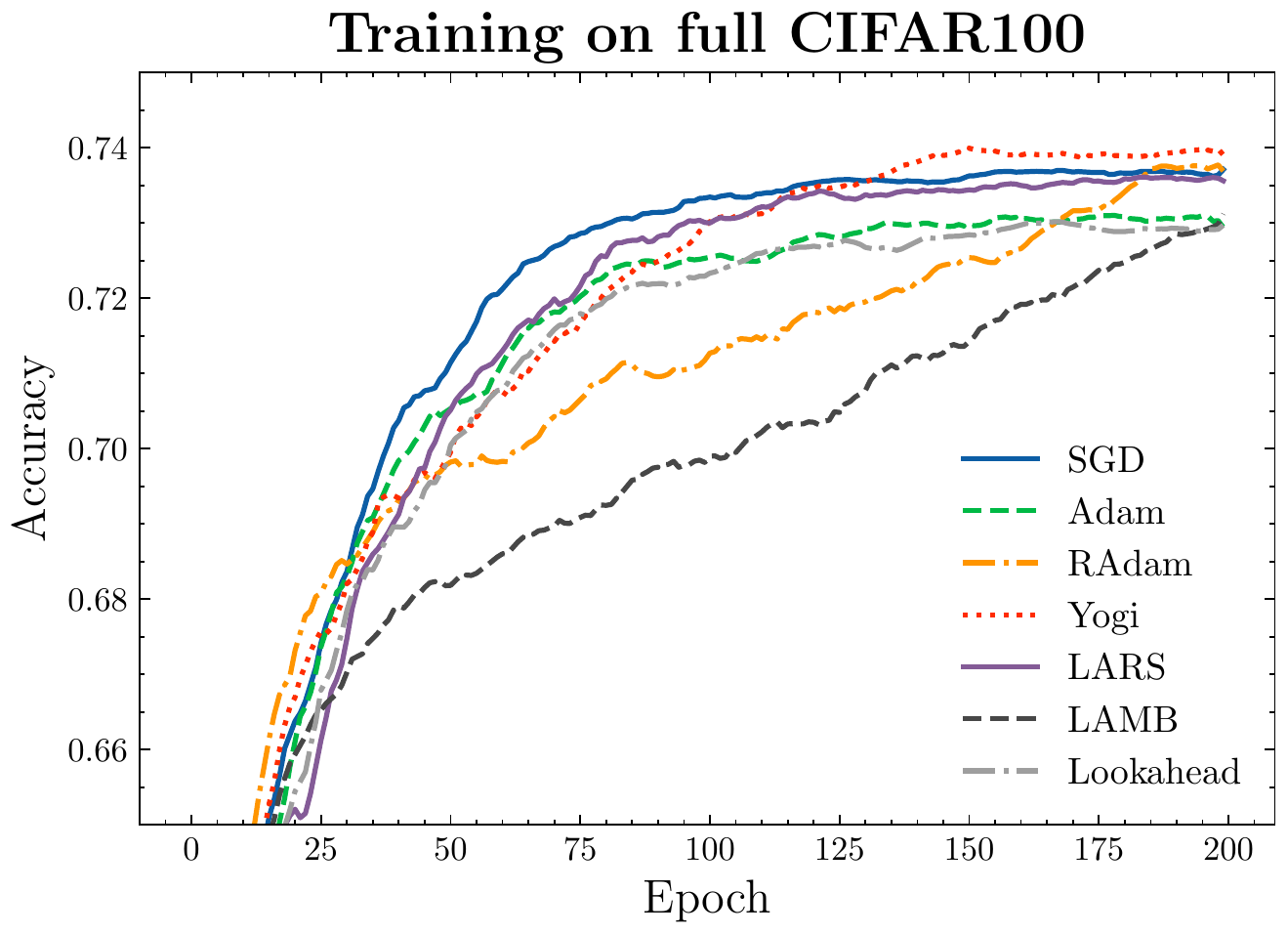}
        \caption{} 
        \label{fig:cifar100-addition}
    \end{subfigure}
    \vspace{-5pt}
    \caption{Training curves under $\Omega_\text{partial}$ for both partial and full datasets.}
    \vspace{-10pt}
    \label{fig:cifar_addition}
\end{figure}

\vspace{-5pt}
We then conduct evaluation on data-addition training based on the protocol in Algorithm \ref{al:data-addition}. We choose four classification problems on CIFAR10, CIFAR100, MRPC and PPI since the setting do not apply to RL. We search the best hyperparameter configuration, denoted by $\Omega_\text{partial}$, under the sub training set with the ratio $\delta=0.3$. Here we tune all hyperparameters. Then we directly apply $\Omega_\text{partial}$ on the full dataset for a complete training process. 
The training curves are shown in Figure \ref{fig:cifar_addition}, and 
we also summarize the training curve with \cpe by Eq.~\ref{eq:tunability} in Table \ref{tab:cpe-addition}. 
We have the following findings: 
\begin{itemize}[noitemsep,topsep=0pt,parsep=0pt,partopsep=0pt,leftmargin=*]
\item There is no clear winner in data-addition training. RAdam is outperforming other optimizers in 2/4 tasks so is slightly preferred, but other optimizers except Lookahead are also competitive (within 1\% range) on at least 2/4 tasks. 
\item To investigate whether the optimizer's ranking will change when adding 70\% data, we compare the training curve on the original 30\% data versus the training curve on the full 100\% data in Figure~\ref{fig:cifar_addition}. We observe that the ranking of optimizers slightly changes after data addition.  
\end{itemize}

\begin{table}[ht]
\caption{CPE of different optimizers computed under curves trained with $\Omega_\text{partial}$ on four full datasets.}
\vspace{-10pt}
\label{tab:cpe-addition}
\begin{center}
   \resizebox{.7\textwidth}{!}{
\begin{tabular}{ccccccccc}
\toprule
{\bf Optimizer}  &\multicolumn{1}{c}{\bf CIFAR10 (\%)$\uparrow$} &\multicolumn{1}{c}{\bf CIFAR100 (\%)$\uparrow$}
&\multicolumn{1}{c}{\bf MRPC$\uparrow$}
&\multicolumn{1}{c}{\bf PPI (\%)$\uparrow$}\\ 
\midrule
SGD & \textbf{90.04} & \color{blue}\textbf{67.91} &  66.62 & 66.83 \\
Adam & \color{blue}\textbf{90.52} & 67.04 & 73.13  & \textbf{70.42} \\
RAdam & \textbf{90.30} & 67.06 & \color{blue}\textbf{79.01}  &   \color{blue}\textbf{70.84}  \\
Yogi &  \textbf{89.63} &  \textbf{67.58} & 68.40   & 67.99 \\
LARS &  \textbf{90.17} & \textbf{67.29}  & 64.43  &  68.40 \\
LAMB &  \textbf{90.51} &  66.13 & \textbf{78.94}  & 70.11 \\
Lookahead & 88.36 &  67.10  & 68.81  & 69.71 \\
\bottomrule
\end{tabular}
} 
\end{center}
\vspace{-10pt}
\end{table}



\vspace{-8pt}\section{Conclusions and Discussions}
\vspace{-8pt}
In conclusion, we found {\bf there is no strong evidence that newly proposed optimizers consistently outperform Adam}, while each of them may be good for some particular tasks. 
In addition to the proposed two evaluation criteria, there could be other factors that affect the practical performance of an optimizer. First, the {\bf memory consumption} is becoming important for training large DNN models. For instance, although Lookahead performs well in certain tasks, it requires more memory than other optimizers, restricting their practical use in some memory constrained applications. Another important criterion is the {\bf scalability} of optimizers. When training with a massively distributed system, optimizing the performance of a large batch regime (e.g., 32K batch size for ImageNet) is important. LARS and LAMB algorithms included in our study are developed for large batch training. We believe it is another important metric for comparing optimizers worth studying further. 

\bibliography{iclr2021_conference}
\bibliographystyle{iclr2021_conference}

\newpage
\appendix
\section{HyperBand}
We present the whole algorithm for Hyperband in Algorithm~\ref{al:hb}, and you can refer to \citet{li2017hyperband} for more details.
\label{appendix: hyperband}
\begin{algorithm}[ht]
    \caption{Hyperband}
    \label{al:hb}
    
    \textbf{Input:} R, $\eta$\\
    \textbf{Initialization:} $s_\text{max}=\lfloor\log_\eta R\rfloor$ , $B=(s_\text{max}+1)R$
    \begin{algorithmic}[1]
    \For{$s\in\{s_\text{max}, s_\text{max}-1, \dots, 0\}$}
    \State $n=\lceil\frac{B}{R}\frac{\eta^s}{(s+1)}\rceil$, $r = R\eta^{-s}$
    \State // begin SuccessiveHalving with $(n,r)$ inner loop
    \State $T=\text{random\_get\_configuration}(n)$
    \For{$i\in\{ 0, \dots, s\}$}
    \State $n_i=\lfloor n \eta^{-i}\rfloor $, $r_i=r\eta^i$
    \State $L=\{\text{run\_then\_return\_val\_loss}(t, r_i)\text{: }t\in T\}$
    \State $T=\text{top\_k}(T, L, \lfloor n_i/\eta \rfloor)$
    \EndFor
    \EndFor
    \end{algorithmic}
    \Return{Hyperparameter configuration with the smallest loss seen so far}
\end{algorithm}

\section{Optimizers}
\textbf{Notations.} Given a vector of parameters $\theta\in\R^d$, we denote a sub-vector of its i-th layer's parameters by $\theta^{(i)}$. $\{\alpha_t\}_{t=1}^{T}$ is a sequence of learning rates during the optimization procedure of a horizon $T$.
$\{\phi_t, \psi_t \}_{t=1}^{T}$ represents a sequence of functions to calculate the first-order and second-order momentum of the gradient $g_t$, which are $m_t$ and $v_t$ respectively at time step $t$. Different optimization algorithms are usually specified by the choice of $\phi(\cdot)$ and $\psi(\cdot)$. $\{r_t\}_{t=1}^{T}$ is an additional sequence of adaptive terms to modify the magnitude of the learning rate in some methods. For algorithms only using the first-order momentum, $\mu$ is the , while $\beta_1$ and $\beta_2$ are coefficients to compute the running averages $m$ and $v$. $\epsilon$ is  a small scalar (e.g., $1\times10^{-8}$) used to prevent division by $0$.

\textbf{Generic optimization framework.} Based on, we further develop a thorough generic optimization framework including an extra adaptive term in Algorithm~\ref{al:general}. The debiasing term used in the original version of Adam is ignored for simplicity. Note that for $\{\alpha_t\}_{t=1}^{T}$, different learning rate scheduling strategies can be adopted and the choice of scheduler is also regarded as a tunable hyperparameter. Without loss of generality, in this paper, we only consider a constant value and a linear decay~\citep{shallue2018measuring} in the following equation, introducing $\gamma$ as a hyperparameter.
\begin{equation*}
    \alpha_t = 
    \begin{cases}
\alpha_0, & \text{constant;}\\
\alpha_0 - (1-\gamma) \alpha_0 \frac{t}{T},& \text{linear decay.}
   \end{cases}
\end{equation*}

With this generic framework, we can summarize several popular optimization methods by explicitly specifying $m_t$, $v_t$ and $r_t$ in Table~\ref{tab:opt}. It should be clarified that Lookahead is an exception of the generic framework. In fact it is more like a high-level mechanism, which can be incorporated with any other optimizer. However, as stated in \citet{zhang2019lookahead}, this optimizer is robust to inner optimization algorithm, $k$, and $\alpha_s$ in Algorithm~\ref{al:look}, we still include Lookahead here with Adam as the base for a more convincing and comprehensive evaluation. We consider Lookahead as a special adaptive method, and tune the same hyperparamters for it as other adaptive optimziers.

\begin{algorithm}[ht]
    \caption{Generic framework of optimization methods}
    \label{al:general}
    
    \textbf{Input:} parameter value $\theta_1$, learning rate with scheduling $\{\alpha_t\}$, sequence of functions $\{ \phi_t, \psi_t, \chi_t \}_{t=1}^T$ to compute $m_t$, $v_t$, and $r_t$ respectively.
    \begin{algorithmic}[1]
    \For{$t = 1$ \textbf{to} $T$}
    \State $g_t = \nabla f_t(\theta_{t})$
    \State $m_t = \phi_t(g_1, \cdots, g_t)$
    \State $v_t = \psi_t(g_1, \cdots, g_t)$
    \State $r_t = \chi_t(\theta_t, m_t, v_t) $
    \State $\theta_{t+1} = \theta_{t} - \alpha_t r_t m_t / \sqrt{v_t}$
    \EndFor
    \end{algorithmic}
\end{algorithm}

\begin{table}[ht]
    \centering
    \small
    \caption{
        A summary of popular optimization algorithms with different choices of $m_t$, $v_t$ and $r_t$.
    }
    \begin{tabular}{c|ccc}
         \toprule
        & $m_t$ & $v_t$ & $r_t$  \\ 
         \midrule
         SGD(M) & $\mu m_{t-1} + g_t$ & $1$ & 1\\
         Adam & $\beta_1 m_{t-1} + (1-\beta_1)g_t$ & $\beta_2 v_{t-1} + (1-\beta_2)g_t^2$ &  1 \\
         RAdam & $\beta_1 m_{t-1} + (1-\beta_1)g_t$ & $\beta_2 v_{t-1} + (1-\beta_2)g_t^2$ &  $\sqrt{\frac{(\rho_t-4)(\rho_t-2)\rho_\infty}{(\rho_\infty-4)(\rho_\infty-2)\rho_t}}$\\
         Yogi & $\beta_1 m_{t-1} + (1-\beta_1)g_t$ & $v_{t-1}-(1-\beta_2)\sign(v_{t-1}-g^2_t)g^2_t$ &  1\\
         LARS & $\mu m_{t-1} + g_t$ & 1 &  
        $\lVert \theta_t^{(i)} \rVert/\lVert m_t^{(i)} \rVert$
         \\
         LAMB & $\beta_1 m_{t-1} + (1-\beta_1)g_t$ & $\beta_2 v_{t-1} + (1-\beta_2)g_t^2$ &  $\frac{\lVert \theta_t^{(i)} \rVert}{\lVert m_t^{(i)}/\sqrt{v_t^{(i)}} \rVert}$\\
         Lookahead* & $\beta_1 m_{t-1} + (1-\beta_1)g_t$ & $\beta_2 v_{t-1} + (1-\beta_2)g_t^2$ &  1\\
         \bottomrule
    \end{tabular}
    \label{tab:opt}
\end{table}
\begin{algorithm}[ht]
    \caption{Lookahead Optimizer}
    \label{al:look}
    
    \textbf{Input:} Initial parameters $\theta_0$, objective function $f$, synchronization period $k$, slow weights step size $\alpha_s$, optimizer $A$
    \begin{algorithmic}[1]
    \For{$t = 1,2,\dots$}
    \State Synchronize parameters $\hat{\theta}_{t,0} \leftarrow \theta_{t-1}$
    \For{$i=1,2,\dots,k$}
    \State Sample minibatch of data $d\in\gD$
    \State $\hat{\theta}_{t,i} \leftarrow \hat{\theta}_{t,i-1} + A(L,\hat{\theta}_{t,i-1}, d)$
    \EndFor
    \State Perform outer update $\theta_t\leftarrow \theta_{t-1}+\alpha_s(\hat{\theta}_{t,k}-\theta_{t-1})$
    \EndFor
    \end{algorithmic}
    \Return{Parameters $\theta$}
\end{algorithm}

\section{Task description}
\label{appendix:task}
We make a concrete description of tasks selected for our optimizer evaluation protocol:
\begin{itemize}[noitemsep,topsep=0pt,parsep=0pt,partopsep=0pt,leftmargin=*]
    \item Image classifcation. For this task, we adopt a ResNet-50~\citep{he2016deep} model on CIFAR10 and CIFAR100 with a batch size of $128$ and the maximum epoch of 200 per trial. 
    \item VAE. We use a vanilla variational autoencoder in \citet{kingma2013auto} with five convolutional and five deconvolutional layers with a latent space of dimension $128$ on CelebA. There are no dropout layers. Trained with a batch size of $144$.
    \item GAN. We train SNGAN with the same network architecture and objective function with spectral normalization for CIFAR10 in \citet{miyato2018spectral}, and the batch size of the generator and the discriminator is $128$ and $64$ respectively.
    \item Natural language processing. In this domain, we finetune RoBERTa-base on MRPC, one of the test suit in GLUE benchmark. For each optimizer, we set the maximal exploration budget to be $800$ epochs. The batch size is 16 sentences. 
    \item Graph learning. Among various graph learning problems, we choose node classification as semi-supervised classification. In GCN training, in there are multiple ways to deal with the neighborhood explosion of stochastic optimizers. We choose Cluster-GCN~\cite{chiang2019cluster} as the backbone to handle neighborhood expansion and PPI as the dataset.
    \item Reinforcement learning. We select Walker2d-v3 from OpenAI Gym (\cite{OpenAI2016}) as our training environment, and PPO (\cite{schulman2017proximal}), implemented by OpenAI SpinningUp (\cite{SpinningUp2018}), as the algorithm that required tuning. We use the same architectures for both action value network $Q$ and the policy network $\pi$. We define 40,000 of environment interactions as one epoch, with a batch size of 4,000. The reward we used is the highest average test reward of an epoch during the training.
    
\end{itemize}

\section{Implementation Details}
\label{appendix:details}
Implementation details of our experiments are provided in this section. Specifically, we give the unified search space for all hyperparamters and their default values in Table~\ref{tab:space}. Note that we tune the learning rate decay factor for image classification tasks when tuning every hyperparamter. For the task on MRPC, $\gamma$ is tuned for all experiments. In other cases, we only tune original hyperparamters without a learning rate scheduler.

In addition, Hyperband parameter values for each task are listed in Table~\ref{tab:hyperband}. These parameters are assigned based on properties of different tasks.
\begin{table}[ht]
    \centering
    \begin{tabular}{ccc}
    \toprule
      \multicolumn{1}{c}{\textbf{Hyperparamter}}   & \textbf{Search space} & \textbf{Default value} \\
      \midrule
        $\alpha_0$ & Log-Uniform$(-8,1)$ & $\times$\\
        $\mu$ & Uniform$[0,1]$ & $0$ for SGD and $0.9$ for LARS\\
        $1-\beta_1$ & Log-Uniform$(-4,0)$ & $0.1$\\
        $1-\beta_2$ & Log-Uniform$(-6,0)$ & $0.001$\\
        $\epsilon$ & Log-Uniform$(-8,1)$ & 1e-8\\
        $\gamma$ & Log-Uniform$(-4,0)$ & $\times$\\
     \bottomrule
    \end{tabular}
    \caption{Hyperparamter search space and default value}
    \label{tab:space}
\end{table}

\begin{table}[ht]
    \centering
    \begin{tabular}{cccc}
    \toprule
      \multirow{2}*{\textbf{Task}} & \multicolumn{3}{c}{\textbf{Hyperband parameter}} \\
      \cline{2-4}
      & \multicolumn{1}{c}{$R$} & $n_c$ & $\eta$\\
    \midrule
 \tabincell{c}{Image\\Classification} & $200$ & $172$ & \multirow{6}*{$\eta=3$}\\
 VAE & $50$ & $62$ & \\
 GAN & $200$ & $172$ & \\
 GLUE benchmark & $10$ & $200$ & \\
 Graph learning & $200$ & $200$ & \\
 RL & $200$ & $172$ & \\
     \bottomrule
    \end{tabular}
    \caption{Hyperband parameters for each task.}
    \label{tab:hyperband}
\end{table}

\section{Additional results}
\label{appendix:results}
More detailed experimental results are reported in this section. 
\subsection{End-to-end training}
\begin{figure}[ht]
    \centering
    \includegraphics[width=0.45\textwidth]{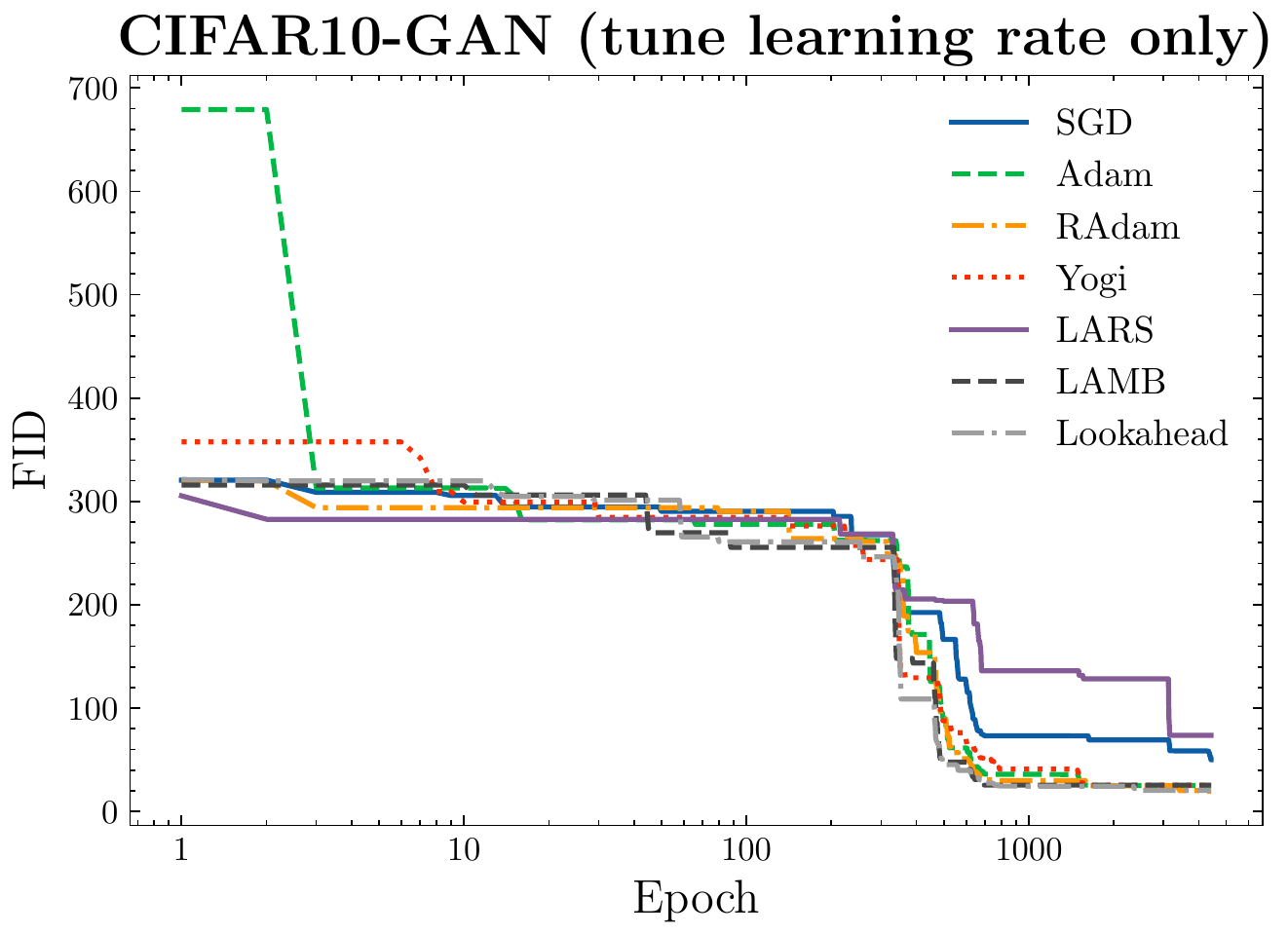}
    \caption{End-to-end training curves for GAN on CIFAR10.}
    \label{fig:gan}
\end{figure}

\begin{figure}[ht] 
    \centering
    \begin{subfigure}[ht]{0.45\textwidth}
        \centering
        \includegraphics[width=\textwidth, 
        trim={0in 0in 0in 0in},
        clip=false]{figures/cifar10-1p.pdf}
        \caption{} 
    \end{subfigure}
    \begin{subfigure}[ht]{0.45\textwidth}
    \centering
    \includegraphics[width=\textwidth, 
    trim={0in 0in 0in 0in},
    clip=false]{figures/cifar10-full.pdf}
    \caption{} 
    \end{subfigure}
    \\
        \centering
    \begin{subfigure}[ht]{0.45\textwidth}
        \centering
        \includegraphics[width=\textwidth, 
        trim={0in 0in 0in 0in},
        clip=false]{figures/cifar100-1p.pdf}
        \caption{} 
    \end{subfigure}
    \begin{subfigure}[ht]{0.45\textwidth}
    \centering
    \includegraphics[width=\textwidth, 
    trim={0in 0in 0in 0in},
    clip=false]{figures/cifar100-full.pdf}
    \caption{} 
    \end{subfigure}
        \centering
    \begin{subfigure}[ht]{0.45\textwidth}
        \centering
        \includegraphics[width=\textwidth, 
        trim={0in 0in 0in 0in},
        clip=false]{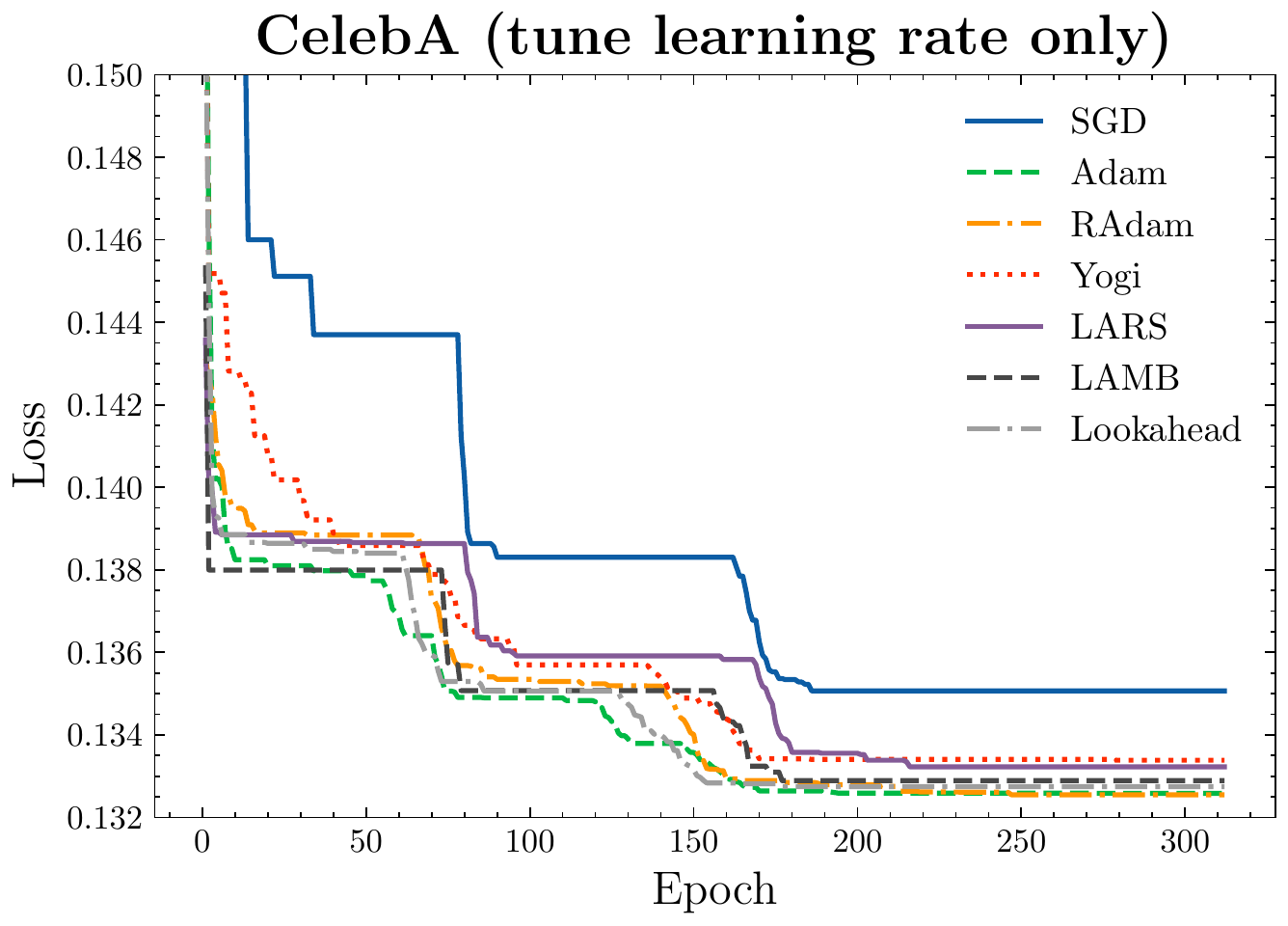}
        \caption{} 
    \end{subfigure}
    \begin{subfigure}[ht]{0.45\textwidth}
    \centering
    \includegraphics[width=\textwidth, 
    trim={0in 0in 0in 0in},
    clip=false]{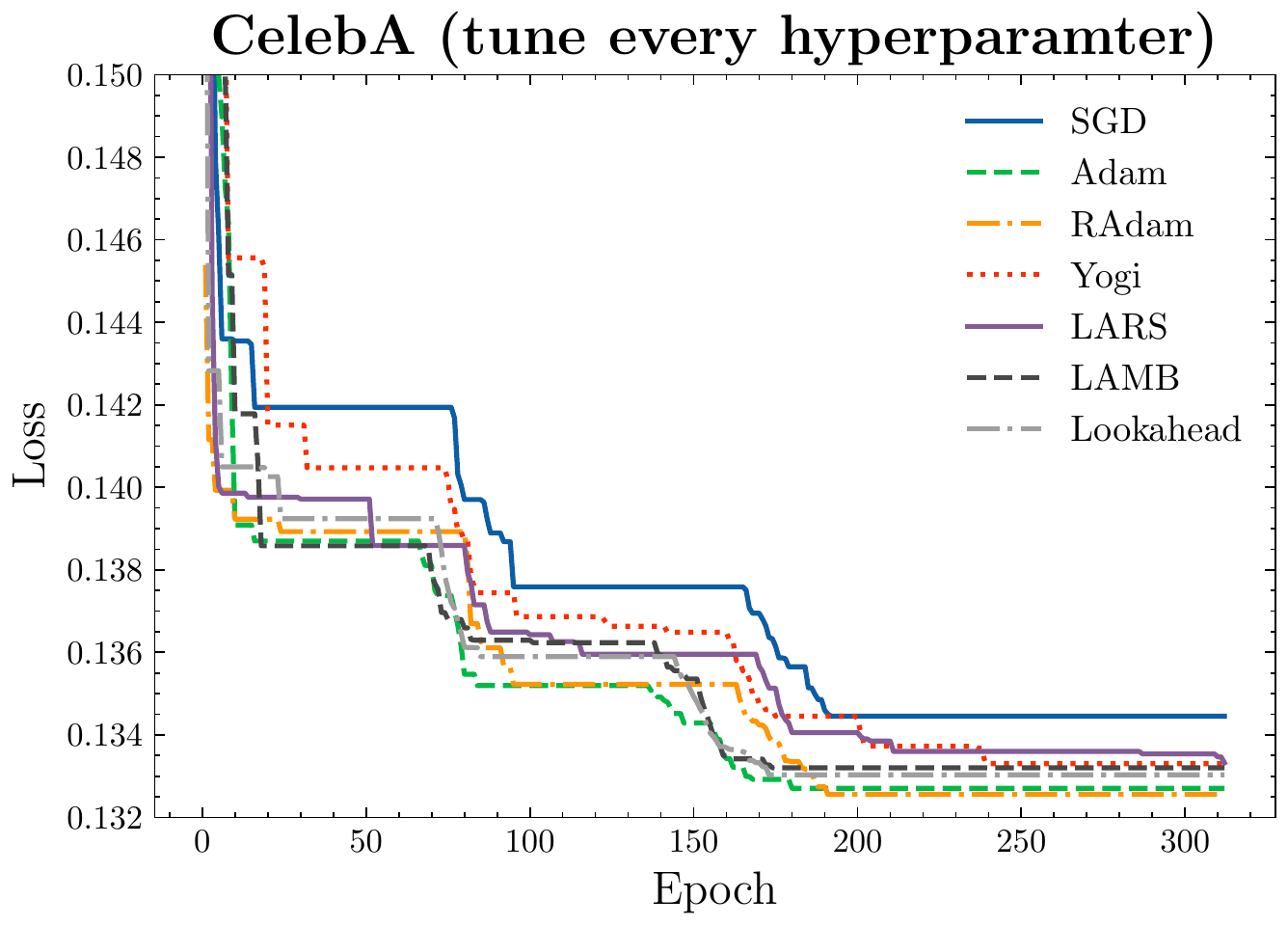}
    \caption{} 
    \end{subfigure}
    \caption{End-to-end training curves on CIFAR100, CIFAR100, and CelebA.}
    \label{fig:ccc}
\end{figure}

\begin{figure}
    \centering
    \begin{subfigure}[ht]{0.45\textwidth}
        \centering
        \includegraphics[width=\textwidth, 
        trim={0in 0in 0in 0in},
        clip=false]{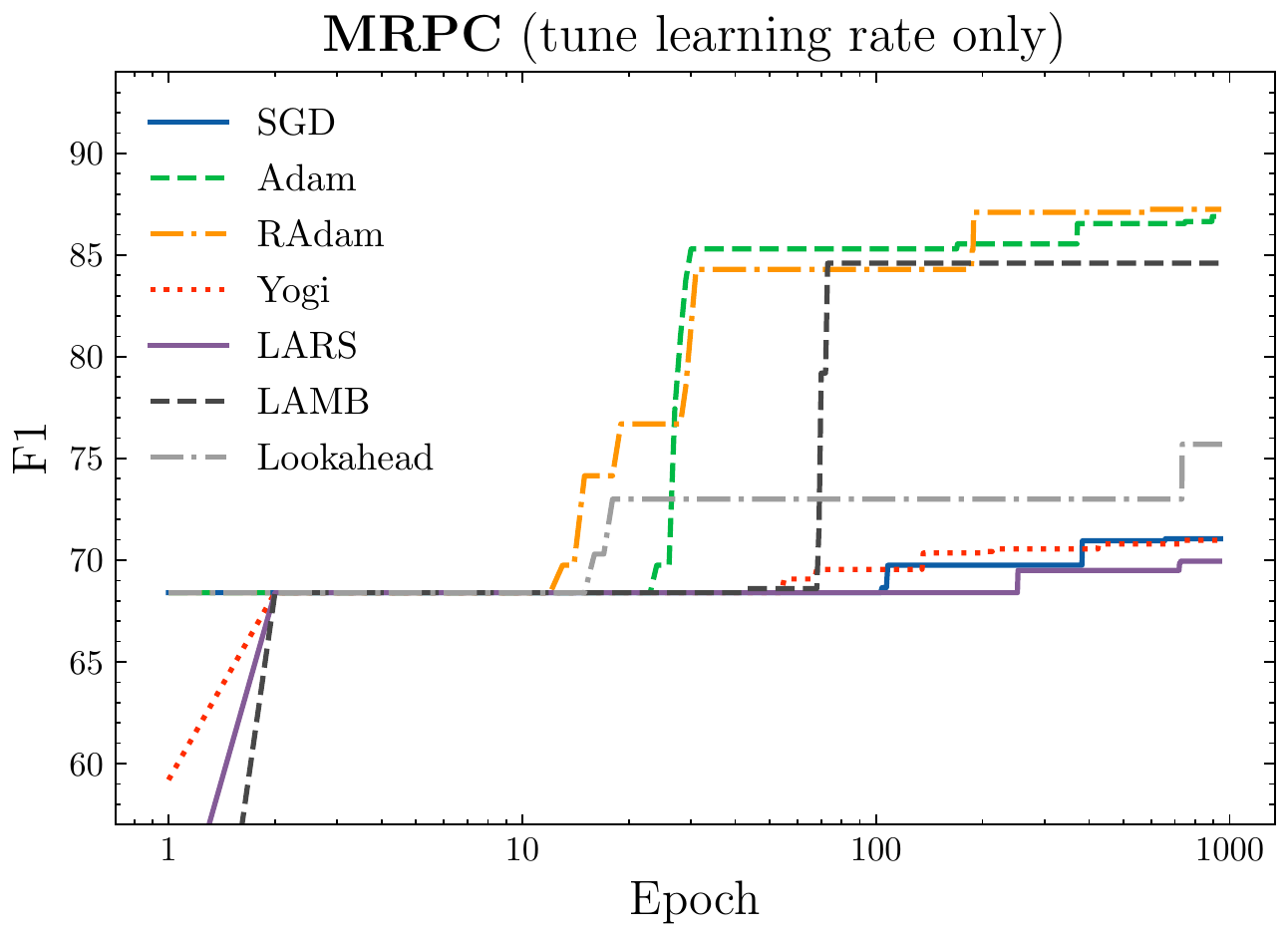}
        \caption{} 
    \end{subfigure}
    \begin{subfigure}[ht]{0.45\textwidth}
    \centering
    \includegraphics[width=\textwidth, 
    trim={0in 0in 0in 0in},
    clip=false]{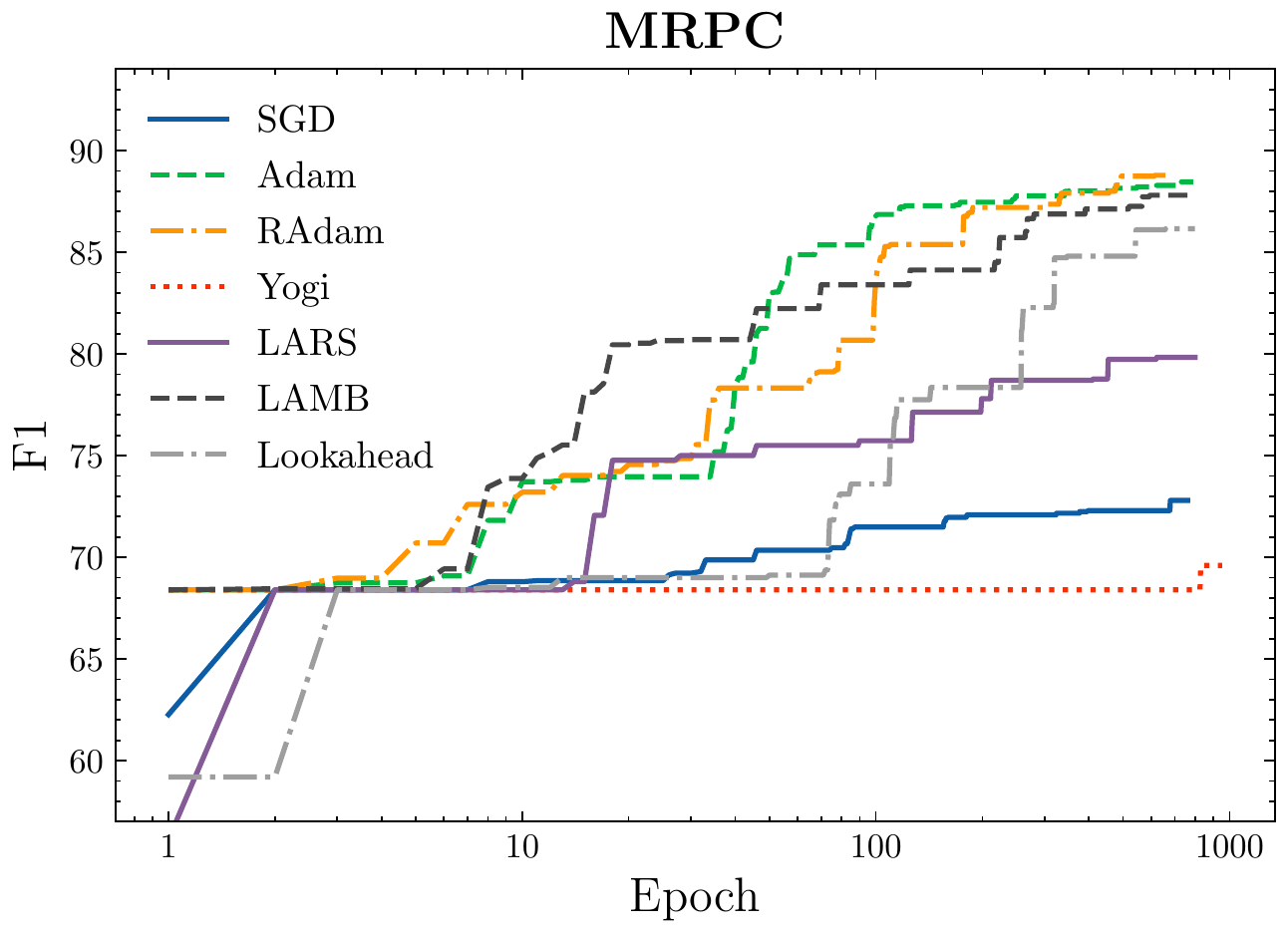}
    \caption{} 
    \end{subfigure}
    \\
        \centering
    \begin{subfigure}[ht]{0.45\textwidth}
        \centering
        \includegraphics[width=\textwidth, 
        trim={0in 0in 0in 0in},
        clip=false]{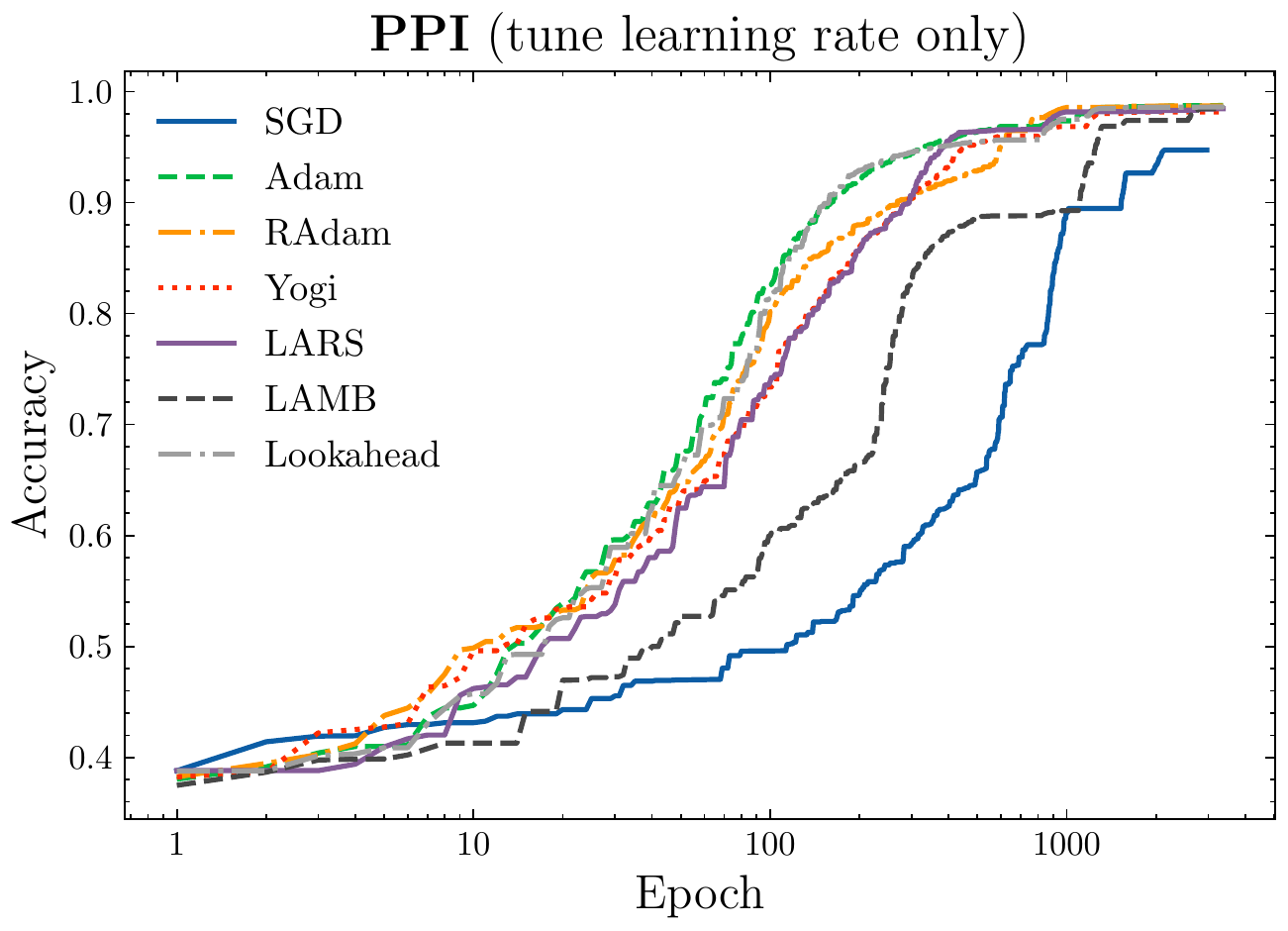}
        \caption{} 
    \end{subfigure}
    \begin{subfigure}[ht]{0.45\textwidth}
    \centering
    \includegraphics[width=\textwidth, 
    trim={0in 0in 0in 0in},
    clip=false]{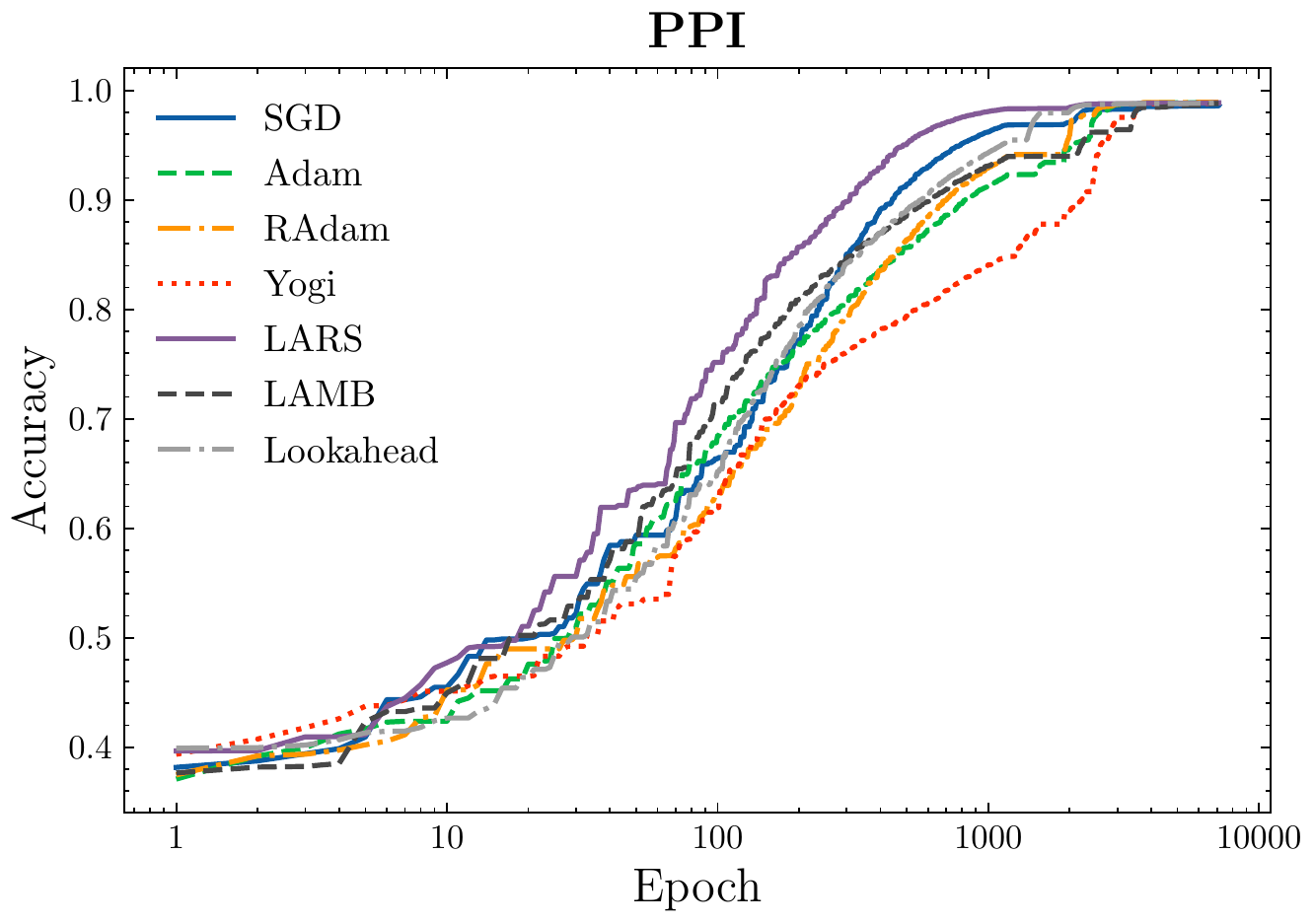}
    \caption{} 
    \end{subfigure}
    \centering
    \begin{subfigure}[ht]{0.45\textwidth}
        \centering
        \includegraphics[width=\textwidth, 
        trim={0in 0in 0in 0in},
        clip=false]{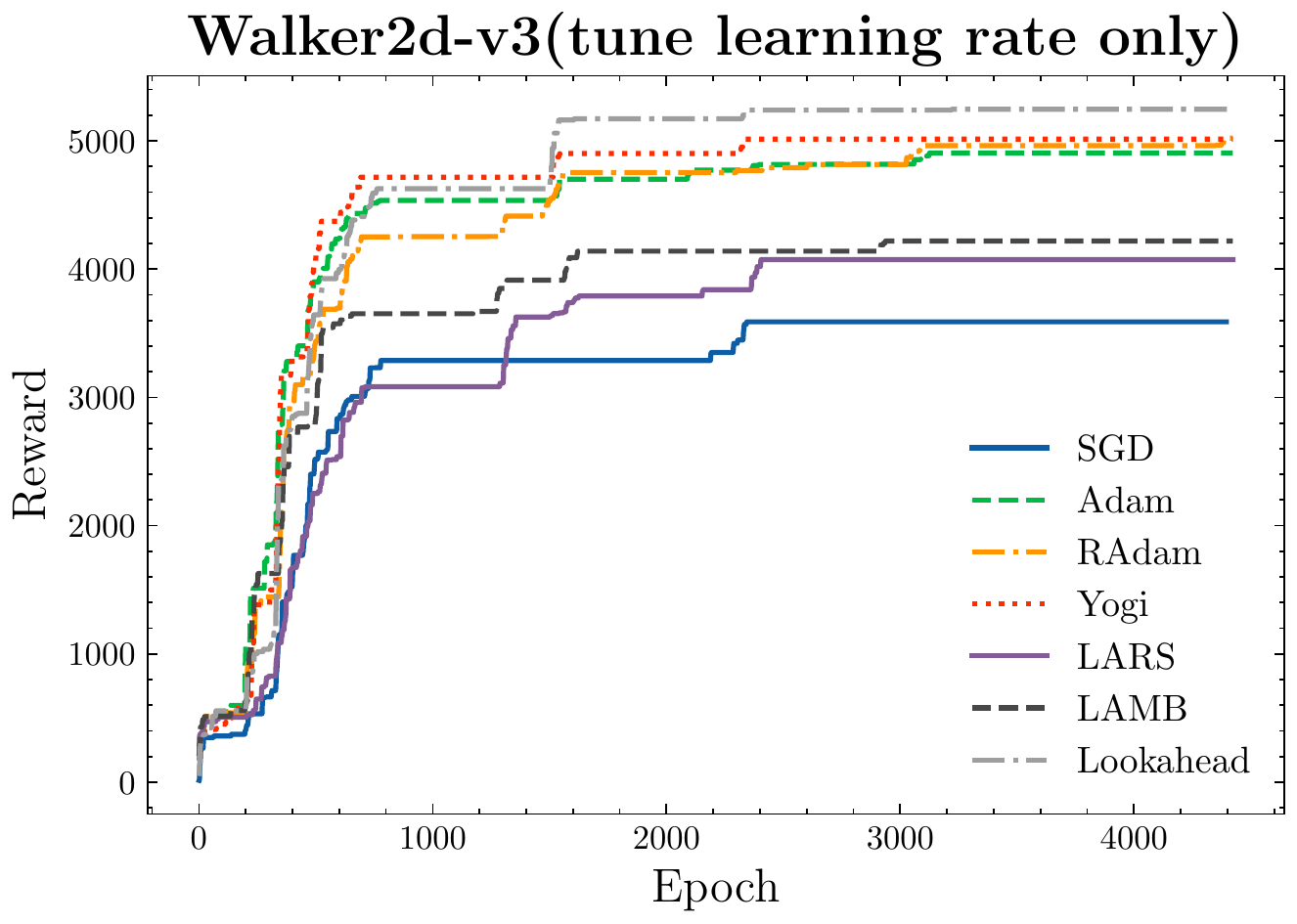}
        \caption{} 
    \end{subfigure}
    \begin{subfigure}[ht]{0.45\textwidth}
    \centering
    \includegraphics[width=\textwidth, 
    trim={0in 0in 0in 0in},
    clip=false]{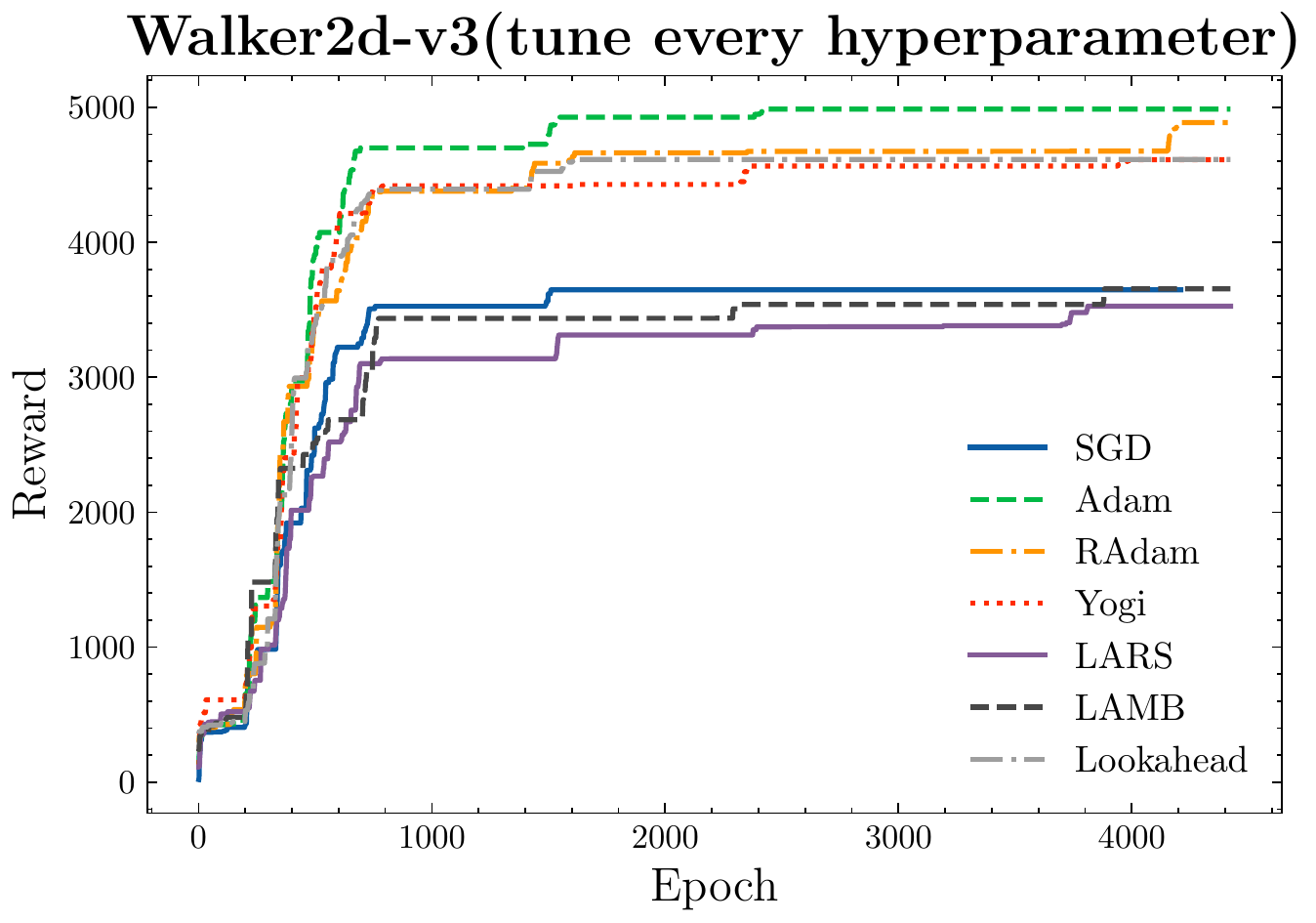}
    \caption{} 
    \end{subfigure}
    \caption{End-to-end training curves on MRPC, PPI, and Walker2d-v3.}
    \label{fig:mpw}
\end{figure}

Table \ref{tab:peak} shows peak performance for optimizers on each task. For GAN, we only conduct evaluation on optimizers tuning learning rate due to time limit, and present its CPE and peak performance in Table~\ref{tab:gan}. There is also an end-to-end training curve for GAN on CIFAR10 in Figure \ref{fig:gan}. Figures for end-to-end training curves on the rest of tasks are shown in Figure~\ref{fig:ccc} and \ref{fig:mpw}.
\begin{table}[ht]
\caption{Peak performance during end-to-end training. The best one for each task is highlighted in bold.
}
\label{tab:peak}
\begin{center}
\scalebox{0.9}{
\begin{tabular}{ccccccc}
\toprule
\multicolumn{1}{c}{\bf Optimizer}  &\multicolumn{1}{c}{\bf CIFAR10 (\%) $\uparrow$} &\multicolumn{1}{c}{\bf CIFAR100 (\%) $\uparrow$}
&\multicolumn{1}{c}{\bf CelebA $\downarrow$}
&\multicolumn{1}{c}{\bf MRPC (\%) $\uparrow$}
&\multicolumn{1}{c}{\bf PPI $\uparrow$}
&\multicolumn{1}{c}{\bf Walker2d-v3 $\uparrow$}
\\ \midrule
\multicolumn{7}{l}{\textit{Tune learning rate only:}}\\
SGD & 93.39 & \textbf{73.68} & 0.1351 & 71.05 & 94.74 & 3589 \\
Adam & 93.65 & 71.51 & 0.1326 & 84.90  &  \textbf{98.73} & 4735\\
RAdam & \textbf{93.93} & 72.30 & \textbf{0.1325}  & \textbf{85.41} &  98.70 & 5020\\
Yogi & 93.58 & 73.48 & 0.1334 & 70.19 & 98.18 & 5013\\
LARS & 93.46 &  73.53 & 0.1332 & 68.97 & 98.45 & 4073\\
LAMB & 93.39 & 70.38 & 0.1329 & 82.23  & 98.46 & 4219\\
Lookahead &  93.60 &71.75 & 0.1327  & 72.99 & 98.63 & \textbf{5246}\\
\midrule
\multicolumn{7}{l}{\textit{Tune every hyperparameter:}}\\
SGD & 93.47 & \textbf{73.94} & 0.1344 & 72.80 & 98.64 & 3647 \\
Adam & 92.58 & 73.82 & 0.1327 & 88.46  & \textbf{98.93}  &  \textbf{4986}\\
RAdam & \textbf{94.47} & 73.50 & \textbf{0.1326}  & \textbf{88.78}  & 98.92  & 4886\\
Yogi & 93.75 & 73.88 & 0.1333 & 69.60 & 98.85 & 4612\\
LARS & 94.22 &  74.08 & 0.1333 & 79.83 & 98.88 & 3526\\
LAMB & 93.88 & 71.31 & 0.1332 & 87.80  & 98.80 & 3654\\
Lookahead &  93.82 & 72.90 & 0.1330  & 86.15  &  98.87 & 4614\\
\bottomrule
\end{tabular}
} 
\end{center}
\end{table}

\begin{table}[ht]
\caption{CPE on GAN for end-to-end training. The value in the bracket is peak performance.
}
\label{tab:gan}
\begin{center}
\begin{tabular}{cc}
\toprule
\multicolumn{1}{c}{\bf Optimizer}   &\multicolumn{1}{c}{\bf GAN-CIFAR10$\downarrow$}
\\ \midrule
\multicolumn{2}{l}{\textit{Tune learning rate only:}}\\
SGD & 113.25 (50.08) \\
Adam & 77.04 (25.14) \\
RAdam & 73.85 (19.61) \\
Yogi & 76.80  (25.36)\\
LARS & 157.71 (73.82)\\
LAMB & 68.47  (25.55)\\
Lookahead & {\color{blue}{\textbf{65.61}}} (\textbf{20.40}) \\
\bottomrule
\end{tabular} 
\end{center}
\end{table}

\subsection{Data addition training}
\begin{figure}
    \centering
    \begin{subfigure}[ht]{0.45\textwidth}
    \centering
    \includegraphics[width=\textwidth, 
    trim={0in 0in 0in 0in},
    clip=false]{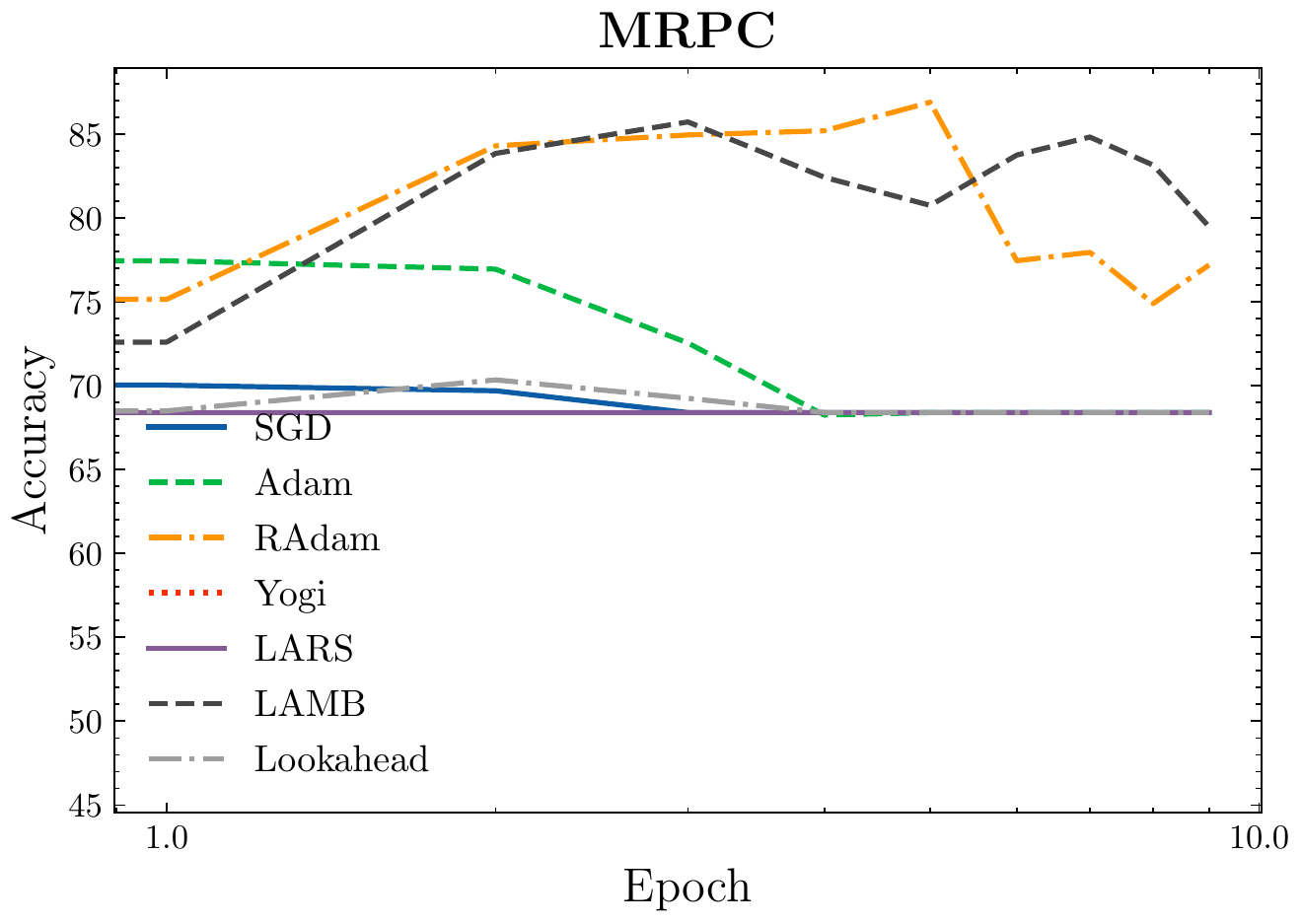}
    \caption{Training on full MRPC.} 
    \end{subfigure}
    \begin{subfigure}[ht]{0.45\textwidth}
    \centering
    \includegraphics[width=\textwidth, 
    trim={0in 0in 0in 0in},
    clip=false]{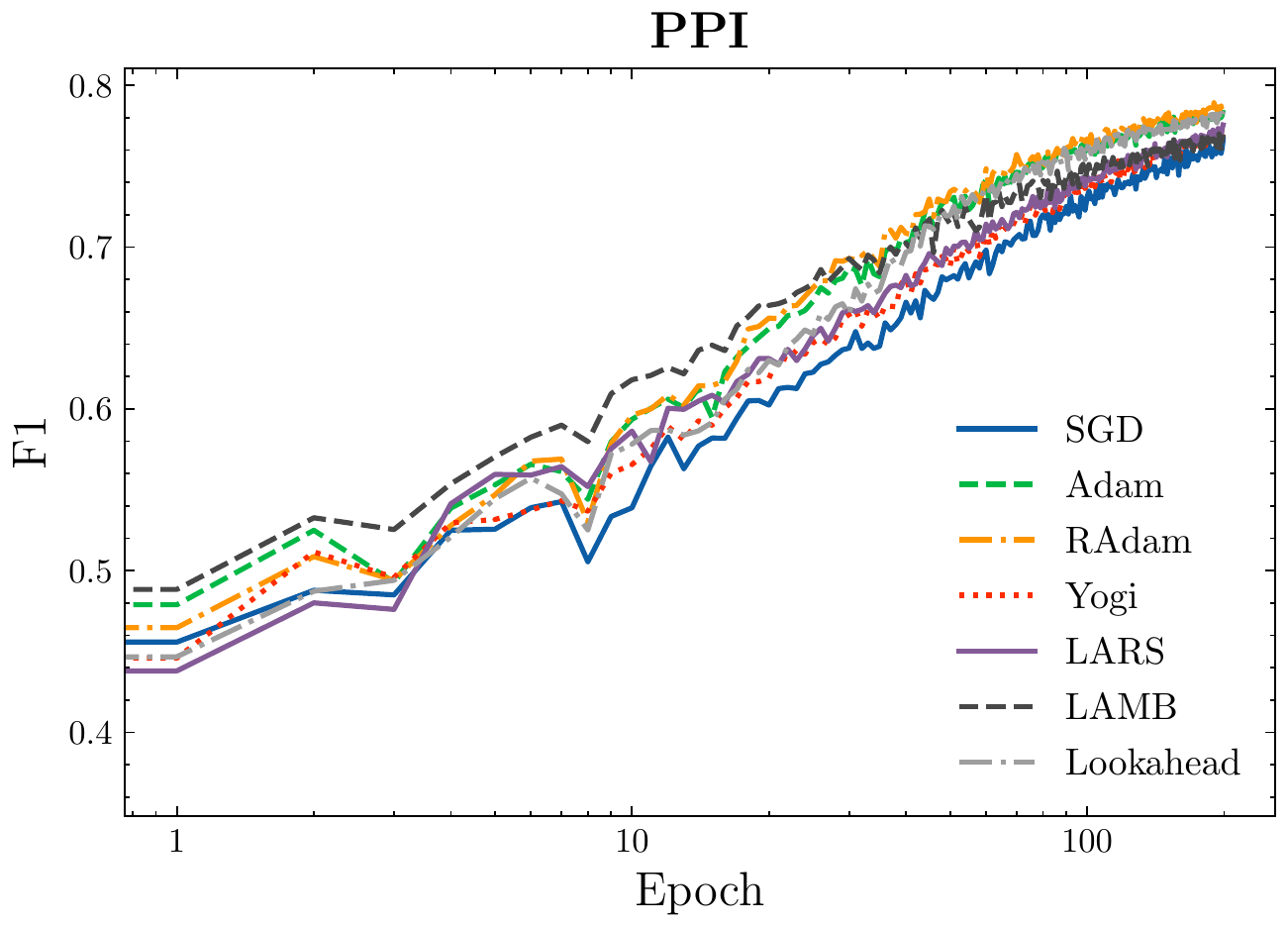}
    \caption{Training on full PPI.} 
    \end{subfigure}
    \caption{Data addition trainong on MRPC and PPI.}
    \label{fig:mrpc-ppi}
\end{figure}

We provide training curves for data-addition training on full MRPC and PPI dataset in Figure~\ref{fig:mrpc-ppi}.

\end{document}